\documentclass{article}

\PassOptionsToPackage{numbers, compress}{natbib}

\usepackage[final]{neurips_2023}

\usepackage[utf8]{inputenc} 
\usepackage[T1]{fontenc}    
\usepackage{hyperref}       
\usepackage{url}            
\usepackage{booktabs}       
\usepackage{amsfonts}       
\usepackage{nicefrac}       
\usepackage{microtype}      
\usepackage{xcolor}         

\usepackage{custom}

\title{Discovering Hierarchical Achievements in Reinforcement Learning via Contrastive Learning}

\author{%
Seungyong Moon$^{1}$, Junyoung Yeom$^{1}$, Bumsoo Park$^{2}$, Hyun Oh Song$^{1}$\thanks{Corresponding author} \\
$^1$Seoul National University, $^2$KRAFTON \\
\texttt{\{symoon11,yeomjy,hyunoh\}@mllab.snu.ac.kr} \\
\texttt{bumsoo.park96@krafton.com} \\
}

\begin{document}

\maketitle

\begin{abstract}

Discovering achievements with a hierarchical structure in procedurally generated environments presents a significant challenge.
This requires an agent to possess a broad range of abilities, including generalization and long-term reasoning.
Many prior methods have been built upon model-based or hierarchical approaches, with the belief that an explicit module for long-term planning would be advantageous for learning hierarchical dependencies.
However, these methods demand an excessive number of environment interactions or large model sizes, limiting their practicality.
In this work, we demonstrate that proximal policy optimization (PPO), a simple yet versatile model-free algorithm, outperforms previous methods when optimized with recent implementation practices.
Moreover, we find that the PPO agent can predict the next achievement to be unlocked to some extent, albeit with limited confidence.
Based on this observation, we introduce a novel contrastive learning method, called \emph{achievement distillation}, which strengthens the agent's ability to predict the next achievement.
Our method exhibits a strong capacity for discovering hierarchical achievements and shows state-of-the-art performance on the challenging Crafter environment in a sample-efficient manner while utilizing fewer model parameters.

\end{abstract}

\section{Introduction}

Deep reinforcement learning (RL) has recently achieved remarkable successes in solving challenging decision-making problems, including video games, board games, and robotic controls \cite{mnih2013playing,silver2016mastering,haarnoja2018soft,schrittwieser2020mastering}.
However, these advancements are often restricted to a single deterministic environment with a narrow set of tasks.
To successfully deploy RL agents in real-world scenarios, which are constantly changing and open-ended, they should generalize well to new unseen situations and acquire reusable skills for solving increasingly complex tasks via long-term reasoning.
Unfortunately, many existing algorithms exhibit limitations in learning these abilities and tend to memorize action sequences rather than truly understand the underlying structures of the environments \citep{machado2018revisiting,kirk2023survey}.

To assess the abilities of agents in generalization and long-term reasoning, we focus on the problem of discovering hierarchical achievements in procedurally generated environments with high-dimensional image observations.
In each episode, an agent navigates a previously unseen environment and receives a sparse reward upon accomplishing a novel subtask labeled as an \textit{achievement}.
Importantly, each achievement is semantically meaningful and can be reused to complete more complex achievements.
Such a setting inherently demands strong generalization and long-term reasoning from the agent.

Previous work on this problem has mainly relied on model-based or hierarchical approaches, which involve explicit modules for long-term planning.
Model-based methods employ a latent world model that predicts future states and rewards for learning long-term dependencies \citep{hafnermastering,hafner2023mastering,anandprocedural,walker2023investigating}.
While these methods have shown effectiveness in discovering hierarchical achievements, particularly in procedurally generated environments, they are constructed with large model sizes and often require substantial exploratory data, which limits their practicality.
Hierarchical methods aim to reconstruct the dependencies between achievements as a graph and employ a high-level planner on the graph to direct a low-level controller toward the next achievement to be unlocked \citep{sohn2020meta,costales2022possibility,zhou2023learning}.
However, these methods rely on prior knowledge of achievements (\eg, the number of achievements), which is impractical in open-world scenarios where the exact number of achievements cannot be predetermined. Additionally, they necessitate a significant number of offline expert data to reconstruct the graph.

To address these issues, we begin by exploring the potential of proximal policy optimization (PPO), a simple and flexible model-free algorithm, in discovering hierarchical achievements \citep{schulman2017proximal}. Surprisingly, PPO outperforms previous model-based and hierarchical methods by adopting recent implementing practices. Furthermore, upon analyzing the latent representations of the PPO agent, we observe that it has a certain degree of predictive ability regarding the next achievement, albeit with high uncertainty.

Based on this observation, we propose a novel self-supervised learning method alongside RL training, named \emph{achievement distillation}. Our method periodically distills relevant information on achievements from episodes collected during policy updates to the encoder via contrastive learning \citep{oord2018representation}. Specifically, we maximize the similarity in the latent space between state-action pairs and the corresponding next achievements within a single episode. Additionally, by leveraging the uniform achievement structure across all environments, we maximize the similarity in the latent space between achievements from two different episodes, matching them using optimal transport \citep{benamou2015iterative}. This learning can be seamlessly integrated into PPO by introducing an auxiliary training phase. Our method demonstrates state-of-the-art performance in discovering hierarchical achievements on the challenging Crafter benchmark, unlocking all 22 achievements with a budget of 1M environment steps while utilizing only 4\% of the model parameters compared to the previous state-of-the-art method \citep{hafner2022benchmarking}.

\section{Preliminaries}

\subsection{Markov decision processes with hierarchical achievements}

We formalize the problem using Markov decision processes (MDPs) with hierarchical achievements \citep{zhou2023learning}.
Let $\mathbb{M}$ represent a collection of such MDPs.
Each environment $\mathcal{M}_i \in \mathbb{M}$ is defined by a tuple $(\mathcal{S}_i, \mathcal{A}, \mathcal{G}, p, r, \rho_i, \gamma)$.
Here, $\mathcal{S}_i \subset \mathcal{S}$ is the image observation space, which has visual variations across different environments, $\mathcal{A}$ is the action space, $\mathcal{G}$ is the achievement graph with a hierarchical structure, $p: \mathcal{S} \times \mathcal{A} \to \mathcal{P}(\mathcal{S})$ is the transition probability function, $r: \mathcal{S} \times \mathcal{A} \times \mathcal{S} \to \mathbb{R}$ is the achievement reward function, $\rho_i \in \mathcal{P}(\mathcal{S}_i)$ is the initial state distribution, and $\gamma \in [0, 1]$ is the discount factor. 

The achievement graph $\mathcal{G} = (\mathcal{V}, \mathcal{E})$ is a directed acyclic graph, where each vertex $v \in \mathcal{V}$ represents an achievement and each edge $(u, v) \in \mathcal{E}$ indicates that achievement $v$ has a dependency on achievement $u$.
To unlock achievement $v$, all of its ancestors (\ie, achievements in the path from the root to $v$) must also be unlocked.
When an agent unlocks a new achievement, it receives an achievement reward of 1.
Note that each achievement can be accomplished multiple times within a single episode, but the agent will only receive a reward when unlocking it for the first time.
Specifically, let $b \in \{0, 1\}^{|\mathcal{V}|}$ be a binary vector indicating which achievements have been unlocked and $c: \mathcal{S} \times \mathcal{A} \times \mathcal{S} \to {\mathcal{V} \cup \{ \emptyset \} }$ be a function determining whether a transition tuple results in the completion of an achievement.
Then, the achievement reward function is defined as
\begin{align*}
r(s_t, a_t, s_{t+1}) = 
\begin{cases}
1 ~~\mathrm{if}~~ \exists v_i \in \mathcal{V}: b[i] = 0, c(s_t, a_t, s_{t+1}) = v_i   \\
0 ~~\mathrm{otherwise}.
\end{cases}
\end{align*}
This reward structure provides an incentive for the agent to explore the environments and discover a new achievement, rather than repeatedly accomplishing the same achievements.

We assume that the agent has no prior knowledge of the achievement graph, including the number of achievements and their dependencies.
Additionally, the agent has no direct access to information about which achievements have been unlocked.
Instead, the agent must infer this information indirectly from the reward signal it receives.
Given this situation, our objective is to learn a generalizable policy $\pi: \mathcal{S} \to \mathcal{P}(\mathcal{A})$ that maximizes the expected return (\ie, unlocks as many achievements as possible) across all environments of $\mathbb{M}$.

\input{figures/crafter_overview}

\subsection{Crafter environment}

We primarily utilize the Crafter environment as a benchmark to assess the capabilities of an agent in solving MDPs with hierarchical achievements \cite{hafner2022benchmarking}.
Crafter is an open-world survival game with 2D visual inputs, drawing inspiration from the popular 3D game Minecraft \cite{guss2019minerldata}.
This is optimized for research purposes, with fast and straightforward environment interactions and clear evaluation metrics.
The game consists of procedurally generated environments with varying world map layouts, terrain types, resource placements, and enemy spawn locations, each of which is uniquely determined by an integer seed.
An agent can only observe its immediate surroundings as depicted in \Cref{fig:crafter_obs}, which makes Crafter partially observable and thus challenging.
To survive, the agent must acquire a variety of skills, including exploring the world map, gathering resources, building tools, placing objects, and defending against enemies.
The game features a set of 22 hierarchical achievements that the agent can unlock by completing specific prerequisites, as illustrated in \Cref{fig:crafter_graph}.
For instance, to make a wood pickaxe, the agent needs to collect wood, place a table, and stand nearby.
This achievement structure is designed to require the agent to learn and utilize a wide range of skills to accomplish increasingly challenging achievements, such as crafting iron tools and collecting diamonds.

\subsection{Proximal policy optimization}

PPO is one of the most successful model-free policy gradient algorithms due to its simplicity and effectiveness \citep{schulman2017proximal}.
PPO learns a policy $\pi_\theta: \mathcal{S} \to \mathcal{P}(\mathcal{A})$ and a value function $V_\theta: \mathcal{S} \to \mathbb{R}$, which are parameterized by neural networks.
During training, PPO first collects a new episodes $\mathcal{T}$ using the policy $\pi_{\theta_\mathrm{old}}$ immediately prior to the update step.
Subsequently, PPO updates the policy network using these episodes for several epochs to maximize the clipped surrogate policy objectives given by
\begin{align*}
J_\pi(\theta) = \mathbb{E}_{(s_t, a_t) \sim \mathcal{T}} \left[ \min \left( \frac{\pi_\theta(a_t \mid s_t)}{\pi_{\theta_\mathrm{old}}(a_t \mid s_t)} \hat{A}_t, \mathrm{clip} \left( \frac{\pi_\theta(a_t \mid s_t)}{\pi_{\theta_\mathrm{old}}(a_t \mid s_t)}, 1 - \epsilon, 1 + \epsilon \right) \hat{A}_t \right) \right],
\end{align*}
where $\hat{A}_t$ is the estimated advantage computed by generalized advantage estimate (GAE) \citep{schulman2016high}.
PPO simultaneously updates the value network to minimize the value objective given by
\begin{align*}
J_V(\theta) = \mathbb{E}_{s_t \sim \mathcal{T}}  \left[ \frac{1}{2} \left( V_\theta(s_t) - \hat{V}_t) \right)^2 \right],
\end{align*}
where $\hat{V}_t = \hat{A}_t + V_{\theta_\mathrm{old}}(s_t)$ is the bootstrapped value function target. 

In image-based RL, it is common practice to optimize the policy and value networks using a shared network architecture \cite{baselines,yarats2021improving}.
An image observation is first passed through a convolutional encoder $\phi_\theta: \mathcal{S} \to \mathbb{R}^h$ to extract a state representation, which is then fed into linear heads to compute the policy and value function.
Sharing state representations between the policy and value networks is crucial to improving the performance of agents in high-dimensional state spaces.
However, relying solely on policy and value optimization to train the encoder can lead to suboptimal state representations, particularly in procedurally generated environments \citep{cobbe2021phasic}. 
To address this issue, recent studies introduce an auxiliary training phase alongside the policy and value optimization that trains the encoder with auxiliary value or self-supervised objectives \citep{cobbe2021phasic,moon2022rethinking}.

\section{Motivation}
\label{sec:motivation}

\subsection{PPO is a strong baseline for hierarchical achievements}

Despite being a simple and ubiquitous algorithm, PPO is less utilized than model-based or hierarchical approaches for solving MDPs with hierarchical achievements.
This is due to the fact that PPO does not have an explicit component for long-term planning or reasoning, which is believed to be essential for solving hierarchical tasks.
However, a recent study has shown that a PPO-based algorithm is also successful in solving hierarchical achievement on the Minecraft environment, albeit with the aid of pre-training on human video data \citep{baker2022video}.

\begin{wrapfigure}{r}{4.6cm}
\vspace{-1em}
\centering
\begin{tikzpicture}
\begin{axis}[
width=5.0cm,
height=5.0cm,
grid=major,
xmin=0,
xmax=1.0e6,
scaled x ticks = false,
tick pos = left,
tick label style={font=\small},
xtick={0, 2e5, 4e5, 6e5, 8e5, 10e5},
xticklabels={0, 0.2, 0.4, 0.6, 0.8, 1.0},
xlabel={Environment steps (M)},
ylabel={Score (\%)},
label style={font=\small},
ylabel near ticks,
xlabel style={font=\small},
ylabel style={font=\small},
legend style={font=\small,at={(0.5, 1.05)},anchor=south,legend columns=1}
]

\addlegendimage{index of colormap=4 of Paired,line width=0.4mm}
\addlegendentry{Size + Layer + Value};

\addlegendimage{index of colormap=0 of Dark2,line width=0.4mm}
\addlegendentry{Size + Layer};

\addlegendimage{index of colormap=4 of Set1,line width=0.4mm}
\addlegendentry{Size};

\addlegendimage{index of colormap=3 of Set1,line width=0.4mm}
\addlegendentry{Default};

\addplot[index of colormap=4 of Paired,line width=0.4mm] table [x=timesteps, y=scores_mean, col sep=comma]{data/ppo_large-score.csv};
\addplot[index of colormap=4 of Paired,name path=plus,draw=none] table [x=timesteps, y=scores_mean_plus_std, col sep=comma]{data/ppo_large-score.csv};
\addplot[index of colormap=4 of Paired,name path=minus,draw=none] table [x=timesteps, y=scores_mean_minus_std, col sep=comma]{data/ppo_large-score.csv};
\addplot[index of colormap=4 of Paired,fill opacity=0.15] fill between[of=plus and minus];

\addplot[index of colormap=0 of Dark2,line width=0.4mm] table [x=timesteps, y=scores_mean, col sep=comma]{data/ppo_old_norm_large-20230506-002141-score.csv};
\addplot[index of colormap=0 of Dark2,name path=plus,draw=none] table [x=timesteps, y=scores_mean_plus_std, col sep=comma]{data/ppo_old_norm_large-20230506-002141-score.csv};
\addplot[index of colormap=0 of Dark2,name path=minus,draw=none] table [x=timesteps, y=scores_mean_minus_std, col sep=comma]{data/ppo_old_norm_large-20230506-002141-score.csv};
\addplot[index of colormap=0 of Dark2,fill opacity=0.15] fill between[of=plus and minus];

\addplot[index of colormap=4 of Set1,line width=0.4mm] table [x=timesteps, y=scores_mean, col sep=comma]{data/ppo_old_large-20230506-001241-score.csv};
\addplot[index of colormap=4 of Set1,name path=plus,draw=none] table [x=timesteps, y=scores_mean_plus_std, col sep=comma]{data/ppo_old_large-20230506-001241-score.csv};
\addplot[index of colormap=4 of Set1,name path=minus,draw=none] table [x=timesteps, y=scores_mean_minus_std, col sep=comma]{data/ppo_old_large-20230506-001241-score.csv};
\addplot[index of colormap=4 of Set1,fill opacity=0.15] fill between[of=plus and minus];

\addplot[index of colormap=3 of Set1,line width=0.4mm] table [x=timesteps, y=scores_mean, col sep=comma]{data/ppo_old_small-20230506-001327-score.csv};
\addplot[index of colormap=3 of Set1,name path=plus,draw=none] table [x=timesteps, y=scores_mean_plus_std, col sep=comma]{data/ppo_old_small-20230506-001327-score.csv};
\addplot[index of colormap=3 of Set1,name path=minus,draw=none] table [x=timesteps, y=scores_mean_minus_std, col sep=comma]{data/ppo_old_small-20230506-001327-score.csv};
\addplot[index of colormap=3 of Set1,fill opacity=0.15] fill between[of=plus and minus];
\end{axis}
\end{tikzpicture}
\caption{Score curves of PPO.}
\label{fig:crafter_ppo}
\vspace{-1em}
\end{wrapfigure}
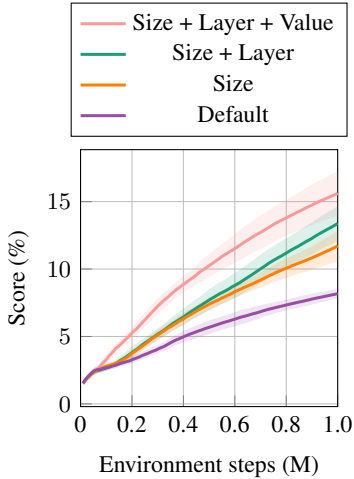

Based on this observation, we first investigate the effectiveness of PPO in solving hierarchical achievements on Crafter without pre-training.
We adopt the recent implementation practices proposed in \citet{andrychowicz2021what,baker2022video}. Concretely, we modify the default ResNet architecture in IMPALA as follows \citep{espeholt2018impala}:
\begin{itemize}[leftmargin=*]
\item \textbf{Network size}: We increase the channel size from [16, 32, 32] to [64, 128, 128] and the hidden size from 256 to 1024. 
\item \textbf{Layer normalization}: We add layer normalization before each dense or convolutional layer \citep{ba2016layer}.
\item \textbf{Value normalization}: We keep a moving average for the mean and standard deviation of the value function targets and update the value network to predict the normalized targets.
\end{itemize}

We train the modified PPO on Crafter for 1M environment steps and evaluate the success rates for unlocking achievements (please refer to \Cref{sec:setup} for the evaluation).
\Cref{fig:crafter_ppo} shows that the slight modification in implementing PPO significantly improves the performance, increasing the score from 8.17 to 15.60.
Notably, this outperforms the current state-of-the-art DreamerV3, which achieves a score of 14.77.

\subsection{Representation analysis of PPO}

Since Crafter environments are procedurally generated, simply memorizing successful episodes is insufficient for achieving high performance.
We hypothesize that the PPO agent acquires knowledge beyond mere memorization of action sequences, possibly including information about achievements.
To validate this, we analyze the learned latent representations of the encoder, as inspired by \citet{wijmans2023emergence}.
Specifically, we collect a batch of episodes using an expert policy and subsample a set of states for training. For each state $s$ in the training set, we freeze its latent representation $\phi_\theta(s)$ from the encoder. 
Subsequently, we train a linear classifier using this representation as input to predict the very next achievement unlocked in the episode containing $s$.
Finally, we evaluate the classification accuracy and the prediction confidence of the ground-truth labels on a held-out test set. The detailed experimental settings are provided in Appendix A.

\begin{wrapfigure}{r}{5.6cm}
\vspace{-1em}
\begin{tikzpicture}
\begin{axis}[
width=6.0cm,
height=5.0cm,
xmin=-0.04,
xmax=1.04,
ymin=0,
ybar,
scaled x ticks = false,
tick pos = left,
tick label style={font=\small},
xlabel={Confidence},
ylabel={Density},
label style={font=\small},
xlabel style={at={(0.5,0.0)}},
ylabel style={at={(0.1,0.5)}},
ytick={0, 1000, 2000, 3000},
yticklabels={0, 0.1, 0.2, 0.3},
legend style={font=\small,at={(0.5, 0.95)},anchor=north,legend columns=2,/tikz/every even column/.append style={column sep=0.2cm}},
legend image code/.code={\draw [#1] (0cm,-0.1cm) rectangle (0.14cm,0.2cm);},
]
\addplot[
index of colormap=4 of Accent,
draw=., 
fill=.,
fill opacity=0.25,
hist={bins=25, data min=0, data max=1.0}
]
table[y=ours, col sep=comma] {data/confidence_new.csv};
\addlegendentry{Ours};
\addplot[
index of colormap=4 of Paired,
draw=., 
fill=.,
fill opacity=0.25,
hist={bins=25, data min=0, data max=1.0}
]
table[y=ppo, col sep=comma] {data/confidence_new.csv};
\addlegendentry{PPO};
\end{axis}
\end{tikzpicture}
\caption{Histogram for the confidence of next achievement prediction.}
\label{fig:confidence}
\vspace{-2em}
\end{wrapfigure}
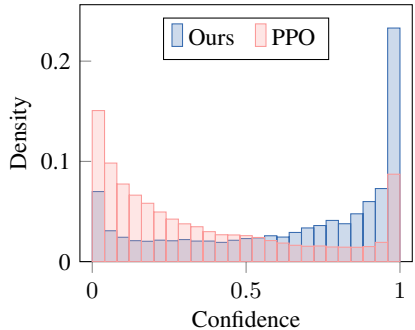

Surprisingly, PPO achieves a nontrivial accuracy of 44.9\% in the 22-way classification. However, \Cref{fig:confidence} shows that the prediction outputs lack confidence with a median value of 0.240.
This suggests that the learned representations of the PPO encoder are not strongly correlated with the next achievement to be unlocked and the agent may struggle to generate optimal action sequences towards a specific goal.

This finding warrants providing additional guidance to the encoder for predicting the next achievements with high confidence.
However, since the agent has no access to the achievement labels, it is challenging to guide the agent in a supervised fashion.
Therefore, it is necessary to explore alternative approaches to guide the agent toward predicting the next achievements.

\section{Contrastive learning for achievement distillation}

In this section, we introduce a new self-supervised learning method that works alongside RL training to guide the encoder in predicting the next achievement to be unlocked. This approach distills relevant information about discovered achievements from episodes collected during multiple policy updates into the encoder via contrastive learning.
This method consists of two key components:
\begin{itemize}[leftmargin=*]
\item \textbf{Intra-trajectory achievement prediction}: Within an episode, this maximizes the similarity in the latent space between a state-action pair and its corresponding next achievement.
\item \textbf{Cross-trajectory achievement matching}: Between episodes, this maximizes the similarity in the latent space for matched achievements.
\end{itemize}

For ease of notation, we denote the sequence of unlocked achievements within an episode as $( g_i )_{i=1}^m$ and their corresponding timesteps as $( t_i )_{i=1}^m$, where each achievement $g_i$ is defined by a transition tuple $(s_{t_i}, a_{t_i}, s_{t_i+1})$. 
For each timestep $t$, we represent the very next achievement as $g_t^+ = g_u$, where $u = \min \{ i \mid t \leq t_i \}$, and the very previous achievement as $g_t^- = g_l$, where $l = \max \{ i \mid t > t_i \}$.

\begin{figure}
\includegraphics[width=\textwidth]{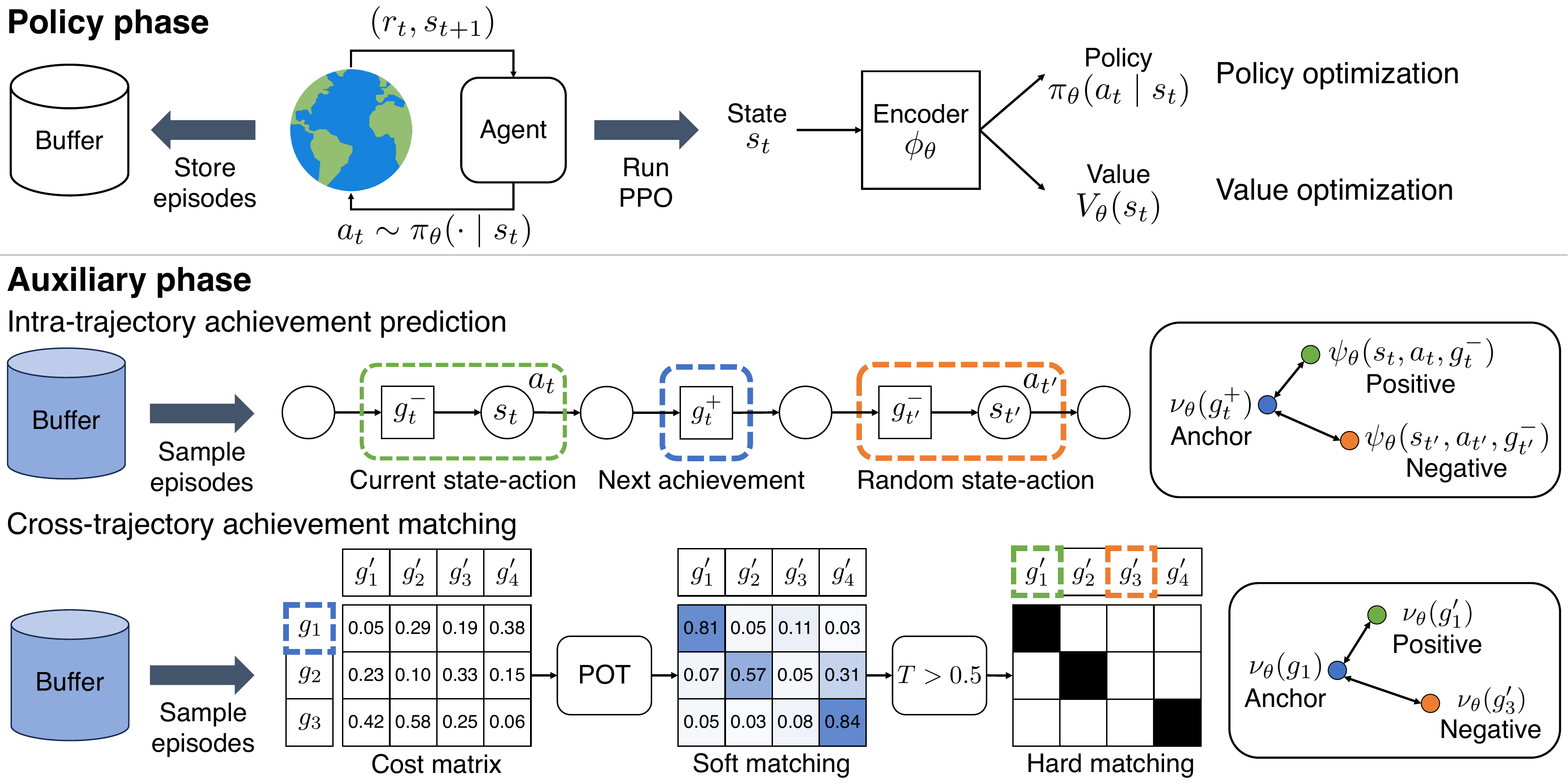}
\caption{Illustration of achievement distillation.}
\label{fig:illustration}
\end{figure}

\subsection{Intra-trajectory achievement prediction}
\label{sec:intra}

Given a state-action pair $(s_t, a_t)$ and its corresponding next achievement $g_t^+$ within an episode $\tau$, we train the encoder $\phi_\theta$ to produce similar representations for them through contrastive learning \citep{oord2018representation,hjelm2018learning}. 
Specifically, we regard $g_t^+$ as the anchor and $(s_t, a_t)$ as the positive.
We also randomly sample another state-action pair $(s_{t'}, a_{t'})$ from the same episode to serve as the negative.
Subsequently, we obtain the normalized representations of the anchor, positive, and negative, denoted as $\nu_\theta(g_t^+)$, $\psi_\theta(s_t, a_t)$, and $\psi_\theta(s_{t'}, a_{t'})$, respectively.
Finally, we minimize the following contrastive loss to maximize the cosine similarity between the anchor and positive representations while minimizing the cosine similarity between the anchor and negative representations:
\begin{align*}
L_\mathrm{pred}(\theta) = - \mathbb{E}_{\substack{(s_t, a_t) \sim \tau \\ (s_{t'}, a_{t'}) \sim \tau}} \left[ \log \left( \frac{\exp(\psi_\theta(s_t, a_t)^\top \nu_\theta(g_t^+) / \lambda) }{\exp (\psi_\theta(s_t, a_t)^\top \nu_\theta(g_t^+) / \lambda) + \exp( \psi_\theta(s_{t'}, a_{t'})^\top \nu_\theta(g_t^+) / \lambda)} \right) \right],
\end{align*}
where $\lambda > 0$ is the temperature parameter.

To obtain the state-action representation $\psi_\theta(s_t, a_t)$, we calculate the latent representation of the state $\phi_\theta(s_t)$ from the encoder and concatenate it with the action $a_t$ using a FiLM layer \citep{perez2018film}. 
The resulting vector is then passed through an MLP layer and normalized.
To obtain the achievement representation $\nu_\theta(g_t^+)$, we simply calculate the residual of the latent representations of the two consecutive states from the encoder and normalize it, as motivated by \citet{nair2019causal}.

\newpage
While this contrastive objective encourages the encoder to predict the next achievement in the latent space, it can potentially lead to distortions in the policy and value networks due to the changes in the encoder. 
To address this, we jointly minimize the following regularizers to preserve the outputs of the policy and value networks, following the practice in \citet{moon2022rethinking}:
\begin{align*}
R_\pi(\theta) = \mathbb{E}_{s_t \sim \tau} \left[ D_\mathrm{KL} \left( \pi_{\theta_\mathrm{old}} (\cdot \mid s_t) \parallel \pi_\theta(\cdot \mid s_t) \right) \right], ~~ R_V(\theta) = \mathbb{E}_{s_t \sim \tau} \left[ \frac{1}{2} \left( V_\theta(s_t) - V_{\theta_\mathrm{old}}(s_t) \right)^2 \right],
\end{align*}
where $\pi_{\theta_\mathrm{old}}$ and $V_{\theta_\mathrm{old}}$ are the policy and value networks immediately prior to the contrastive learning, respectively and $D_\mathrm{KL}$ denotes the KL divergence.

\subsection{Cross-trajectory achievement matching}
\label{sec:cross}

Since Crafter environments are procedurally generated, the achievement representations learned solely from intra-trajectory information may include environment-specific features that limit generalization.
To obtain better achievement representations, we leverage the common achievement structure shared across all episodes.

We first match the sequences of unlocked achievements from two different episodes in an unsupervised fashion.
Given two achievement sequences $\mathbf{g}=( g_i )_{i=1}^m$ and $\mathbf{g}' = ( g_j' )_{j=1}^n$, we define the cost matrix $M \in \mathbb{R}^{m \times n}$ as the cosine distance between the achievement representations:
\begin{align*}
M_{ij} = 1 - \nu_\theta(g_i)^\top \nu_\theta(g_j').
\end{align*}
Subsequently, we regard these two sequences as discrete uniform distributions and compute a soft-matching $T \in \mathbb{R}^{m \times n}$ between them using partial optimal transport, which can be solved by
\begin{align*}
T &= \underset{T \geq 0}{\arg\min} ~ \langle T, M \rangle + \alpha \sum_{i=1}^m \sum_{j=1}^n T_{ij} \log T_{ij} \\
&\quad\,\, \mathrm{subject~to} ~~ T \mathbf{1} \leq \mathbf{1}, ~ T^\top \mathbf{1} \leq \mathbf{1},  ~ \mathbf{1}^\top T^\top \mathbf{1} = \min \{ m, n \},
\end{align*}
where $\alpha > 0$ is the entropic regularization parameter \citep{benamou2015iterative}.
Here, we set the total amount of probability mass to be transported to the minimum length of the two sequences for simplicity.
However, some unlocked achievements in one sequence may not exist in the other sequence, and therefore should not be transported. 
In this case, the optimal amount of probability mass to be transported should be less than the minimum length.
To address this, we compute the conservative hard matching $T^*$ from $T$ by thresholding the probabilities less than 0.5 (\ie, $T^* = \mathbbm{1} [T > 0.5]$).
Note that this also encourages each achievement to be matched to at most one other achievement. We provide examples of matching results in Appendix B.

We train the encoder to produce similar representations for the matched achievements according to $T^*$ through contrastive learning.
Specifically, suppose that the $i$th achievement of the source sequence $g_i$ is matched with the $k$th achievement of the target sequence $g_k'$.
Then, we consider $g_i$ as the anchor and $g_k'$ as the positive. 
We also randomly sample another achievement $g_j'$ from the target sequence to serve as the negative. 
Subsequently, we obtain the normalized representations of these achievements. Finally, we minimize the following contrastive loss to maximize the cosine similarity between the anchor and positive representations while minimizing the cosine similarity between the anchor and negative representations:
\begin{align*}
L_\mathrm{match}(\theta) = - \mathbb{E}_{g_i \sim \mathbf{g},  g'_j \sim \mathbf{g}'} \left[ \log \left( \frac{\exp(\nu_\theta(g_i)^\top \nu_\theta(g_k') / \lambda) }{\exp (\nu_\theta(g_i)^\top \nu_\theta(g_k') / \lambda) + \exp( \nu_\theta(g_i)^\top \nu_\theta(g'_j) / \lambda)} \right) \right],
\end{align*}
As in \Cref{sec:intra}, we jointly minimize the policy and value regularizers to prevent distortions.

\subsection{Achievement representation as memory}
\label{sec:mem}

We further utilize the achievement representations learned in \Cref{sec:intra,sec:cross} as memory for the policy and value networks.
Specifically, given a state $s_t$ and its corresponding previous achievement $g_t^-$, we concatenate the latent state representation $\phi_\theta(s_t)$ with the previous achievement representation $\nu_\theta(g_t^-)$. The resulting vector is then fed into the policy and value heads to output an action distribution and value estimate, respectively.

We also utilize the previous achievement for the achievement prediction task in \Cref{sec:intra}. Given a state-action pair $(s_t, a_t)$ and its corresponding previous achievement $g_t^-$, we concatenate the latent state representation $\phi_\theta(s_t)$ from the encoder with $a_t$ and $\nu_\theta(g_t^-)$. The resulting vector is then fed into an MLP layer and normalized to obtain the representation $\psi_\theta(s_t, a_t, g_t^-)$ for the next achievement prediction. This is interpreted as learning forward dynamics in the achievement space.

\subsection{Integration with RL training}

We integrate the contrastive learning method proposed in \Cref{sec:intra,sec:cross,sec:mem} with PPO training by introducing two alternating phases, the policy and auxiliary phases.
During the policy phase, which is repeated several times, we update the policy and value networks using newly-collected episodes and store them in a buffer.
During the auxiliary phase, we update the encoder to optimize the contrastive objectives in conjunction with the policy and value regularizers using all episodes in the buffer.
We call this auxiliary learning \emph{achievement distillation}. The illustration and pseudocode are presented in \Cref{fig:illustration} and \Cref{alg:achievement}, respectively.

\begin{algorithm}[h]
\caption{PPO with achievement distillation}
\begin{algorithmic}[1]
\Require Policy network $\pi_\theta$, value network $V_\theta$
\For{phase = $1, 2, \ldots$}
\State Reset the buffer $\mathcal{B}$
\For{iter = $1, 2, \ldots, N_\pi$} \Comment{PPO training}
\State Collect episodes $\mathcal{T}$ using $\pi_\theta$ and add them to $\mathcal{B}$
\For{epoch = $1, 2, \ldots, E_\pi$}
\State Optimize $J_\pi(\theta)$ and $J_V(\theta)$ using $\mathcal{T}$
\EndFor
\EndFor
\State $\pi_{\theta_\mathrm{old}} \leftarrow \pi_\theta$, $V_{\theta_\mathrm{old}} \leftarrow V_\theta$
\For{iter = $1, 2, \ldots, E_\mathrm{aux}$}  \Comment{Achievement distillation}
\State Optimize $L_\mathrm{pred}(\theta)$,  $R_\pi(\theta)$, and $R_V(\theta)$ using $\mathcal{B}$
\State Optimize $L_\mathrm{match}(\theta)$,  $R_\pi(\theta)$, and $R_V(\theta)$ using $\mathcal{B}$
\EndFor
\EndFor
\end{algorithmic}
\label{alg:achievement}
\end{algorithm}

\section{Experiments}

\subsection{Experimental setup}
\label{sec:setup}

To assess the effectiveness of our method in discovering hierarchical achievements, we train the agent on Crafter for 1M environment steps and evaluate its performance, following the protocol in \citet{hafner2022benchmarking}. 
We measure the success rates for all 22 achievements across all training episodes as a percentage and calculate their geometric mean to obtain our primary evaluation score\footnote{The score is computed by $S = \exp(\frac{1}{N} \sum_{i=1}^{N} \ln(1 + s_i)) - 1$, where $s_i \in [0, 100]$ is the success rate of the $i$th achievement and $N=22$.}.
Note that the geometric mean prioritizes unlocking challenging achievements.
We also measure the episode reward, which indicates the number of achievements unlocked within a single episode, and report the average across all episodes within the most recent 100K environment steps.
We conduct 10 independent runs using different random seeds for each experimental setting and report the mean and standard deviation.
The code can be found at \url{https://github.com/snu-mllab/Achievement-Distillation}.

We compare our method with our backbone algorithm PPO and four baseline methods that have been previously evaluated on Crafter in other research work: DreamerV3, LSTM-SPCNN, MuZero + SPR, and SEA \citep{hafner2023mastering,stanic2022learning,walker2023investigating,zhou2023learning}. 
DreamerV3 is a model-based algorithm that has achieved state-of-the-art performance on Crafter without any pre-training.
LSTM-SPCNN is a model-free algorithm based on PPO that employs a recurrent and object-centric network to improve performance on Crafter.
MuZero + SPR is a model-based algorithm that has demonstrated state-of-the-art performance on Crafter by utilizing unsupervised pre-training.
SEA is a hierarchical algorithm based on IMPALA that employs a high-level planner to discover achievements on Crafter.
We provide the implementation details and hyperparameters in Appendix C.

\subsection{Crafter results}

\begin{figure}[t]
\centering
\begin{tikzpicture}
\begin{groupplot}[
group style = {group size = 2 by 1, horizontal sep=1.5cm},
width=5.5cm,
height=5.0cm,
grid=major,
xmin=0,
xmax=1.0e6,
scaled x ticks = false,
tick pos = left,
tick label style={font=\small},
xtick={0, 2e5, 4e5, 6e5, 8e5, 10e5},
xticklabels={0, 0.2, 0.4, 0.6, 0.8, 1.0},
xlabel={Environment steps (M)},
ylabel near ticks,
xlabel style={font=\small},
ylabel style={font=\small},
]
\nextgroupplot[
ylabel=Score (\%),
no markers,
legend style={font=\small,legend columns=1},
legend to name=exp,
ytick={0, 5, 10, 15, 20, 25},
ymin=0.0,
ymax=25,
]

\addlegendimage{index of colormap=4 of Accent,line width=0.4mm}
\addlegendentry{Ours};

\addlegendimage{index of colormap=4 of Paired,line width=0.4mm}
\addlegendentry{PPO};

\addlegendimage{index of colormap=0 of Accent,line width=0.4mm}
\addlegendentry{DreamerV3};

\addlegendimage{index of colormap=6 of Paired,line width=0.4mm}
\addlegendentry{LSTM-SPCNN};

\addplot[index of colormap=4 of Accent,line width=0.4mm] table [x=timesteps, y=scores_mean, col sep=comma]{data/ppo_infonce_matching_v7_large-20230505-130607-score.csv};
\addplot[index of colormap=4 of Accent,name path=plus,draw=none] table [x=timesteps, y=scores_mean_plus_std, col sep=comma]{data/ppo_infonce_matching_v7_large-20230505-130607-score.csv};
\addplot[index of colormap=4 of Accent,name path=minus,draw=none] table [x=timesteps, y=scores_mean_minus_std, col sep=comma]{data/ppo_infonce_matching_v7_large-20230505-130607-score.csv};
\addplot[index of colormap=4 of Accent,fill opacity=0.15] fill between[of=plus and minus];

\addplot[index of colormap=4 of Paired,line width=0.4mm] table [x=timesteps, y=scores_mean, col sep=comma]{data/ppo_large-score.csv};
\addplot[index of colormap=4 of Paired,name path=plus,draw=none] table [x=timesteps, y=scores_mean_plus_std, col sep=comma]{data/ppo_large-score.csv};
\addplot[index of colormap=4 of Paired,name path=minus,draw=none] table [x=timesteps, y=scores_mean_minus_std, col sep=comma]{data/ppo_large-score.csv};
\addplot[index of colormap=4 of Paired,fill opacity=0.15] fill between[of=plus and minus];

\addplot[index of colormap=0 of Accent,line width=0.4mm] table [x=timesteps, y=scores_mean, col sep=comma]{data/dreamer-v3-score.csv};
\addplot[index of colormap=0 of Accent,name path=plus,draw=none] table [x=timesteps, y=scores_mean_plus_std, col sep=comma]{data/dreamer-v3-score.csv};
\addplot[index of colormap=0 of Accent,name path=minus,draw=none] table [x=timesteps, y=scores_mean_minus_std, col sep=comma]{data/dreamer-v3-score.csv};
\addplot[index of colormap=0 of Accent,fill opacity=0.15] fill between[of=plus and minus];

\addplot[index of colormap=6 of Paired,line width=0.4mm] table [x=timesteps, y=scores_mean, col sep=comma]{data/lstm-spcnn-score.csv};
\addplot[index of colormap=6 of Paired,name path=plus,draw=none] table [x=timesteps, y=scores_mean_plus_std, col sep=comma]{data/lstm-spcnn-score.csv};
\addplot[index of colormap=6 of Paired,name path=minus,draw=none] table [x=timesteps, y=scores_mean_minus_std, col sep=comma]{data/lstm-spcnn-score.csv};
\addplot[index of colormap=6 of Paired,fill opacity=0.15] fill between[of=plus and minus];

\nextgroupplot[
ylabel=Reward,
no markers,
ytick={0, 4, 8, 12, 16},
ymin=0,
ymax=16,
]

\addplot[index of colormap=4 of Accent,line width=0.4mm] table [x=timesteps, y=rewards_mean, col sep=comma]{data/ppo_infonce_intr_v2_large-20230504-002546-score.csv};
\addplot[index of colormap=4 of Accent,name path=plus,draw=none] table [x=timesteps, y=rewards_mean_plus_std, col sep=comma]{data/ppo_infonce_intr_v2_large-20230504-002546-score.csv};
\addplot[index of colormap=4 of Accent,name path=minus,draw=none] table [x=timesteps, y=rewards_mean_minus_std, col sep=comma]{data/ppo_infonce_intr_v2_large-20230504-002546-score.csv};
\addplot[index of colormap=4 of Accent,fill opacity=0.15] fill between[of=plus and minus];

\addplot[index of colormap=4 of Paired,line width=0.4mm] table [x=timesteps, y=rewards_mean, col sep=comma]{data/ppo_large-score.csv};
\addplot[index of colormap=4 of Paired,name path=plus,draw=none] table [x=timesteps, y=rewards_mean_plus_std, col sep=comma]{data/ppo_large-score.csv};
\addplot[index of colormap=4 of Paired,name path=minus,draw=none] table [x=timesteps, y=rewards_mean_minus_std, col sep=comma]{data/ppo_large-score.csv};
\addplot[index of colormap=4 of Paired,fill opacity=0.15] fill between[of=plus and minus];

\addplot[index of colormap=0 of Accent,line width=0.4mm] table [x=timesteps, y=rewards_mean, col sep=comma]{data/dreamer-v3-score.csv};
\addplot[index of colormap=0 of Accent,name path=plus,draw=none] table [x=timesteps, y=rewards_mean_plus_std, col sep=comma]{data/dreamer-v3-score.csv};
\addplot[index of colormap=0 of Accent,name path=minus,draw=none] table [x=timesteps, y=rewards_mean_minus_std, col sep=comma]{data/dreamer-v3-score.csv};
\addplot[index of colormap=0 of Accent,fill opacity=0.15] fill between[of=plus and minus];

\addplot[index of colormap=6 of Paired,line width=0.4mm] table [x=timesteps, y=rewards_mean, col sep=comma]{data/lstm-spcnn-score.csv};
\addplot[index of colormap=6 of Paired,name path=plus,draw=none] table [x=timesteps, y=rewards_mean_plus_std, col sep=comma]{data/lstm-spcnn-score.csv};
\addplot[index of colormap=6 of Paired,name path=minus,draw=none] table [x=timesteps, y=rewards_mean_minus_std, col sep=comma]{data/lstm-spcnn-score.csv};
\addplot[index of colormap=6 of Paired,fill opacity=0.15] fill between[of=plus and minus];

\end{groupplot}
\node[below=0.09cm, anchor=north west] at (current bounding box.north east) {\pgfplotslegendfromname{exp}};

\end{tikzpicture}
\caption{Score and reward curves.}
\label{fig:crafter_score}
\end{figure}
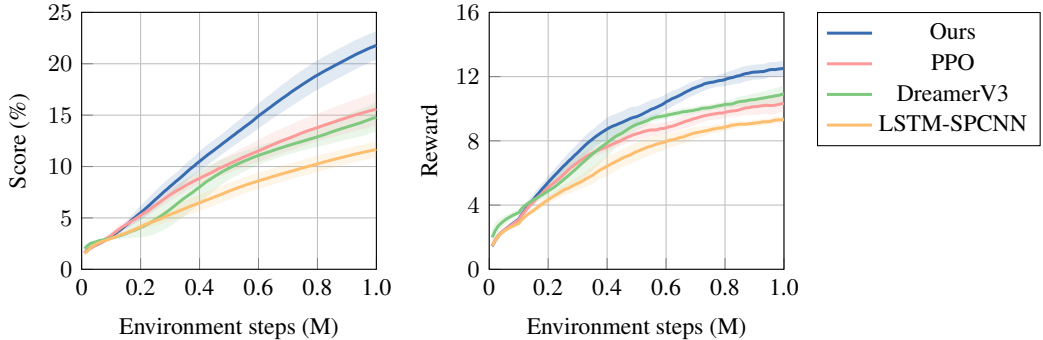

\Cref{tab:crafter_score} and \Cref{fig:crafter_score} present the Crafter scores and rewards obtained by our method and the baselines.
Our method outperforms all the baselines trained from scratch in both the metrics by a considerable margin, with a score of 21.79\% and a reward of 12.60.
Notably, our method exhibits superior score performance compared to MuZero + SPR, which utilizes pre-collected exploratory data and employs a computationally expensive tree search algorithm for planning, while achieving comparable rewards.

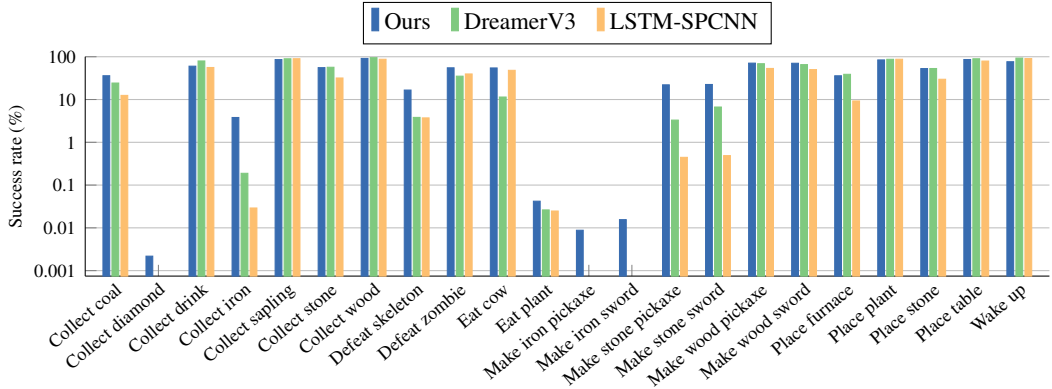
\begin{figure}[t]
\centering
\pgfplotstableread[col sep=comma]{data/dreamer-v3-success_rate.csv}\dreamer
\pgfplotstableread[col sep=comma]{data/ppo_infonce_matching_v7_large-20230505-130607-success_rate.csv}\ours
\pgfplotstableread[col sep=comma]{data/lstm-spcnn-success_rate.csv}\lstm

\begin{tikzpicture}
\begin{axis}[
width=14.4cm,
height=4.5cm,
ymajorgrids=true,
xmin=-0.7,
xmax=21.7,
ymax=100,
axis x line*= bottom,
axis y line*= left,
ymode=log,
log origin=infty,
tick pos=left,
xtick={0,...,21},
xticklabels from table = {\dreamer}{task},
xticklabel style={font=\scriptsize, anchor=north east , rotate=40, xshift=0.2em, yshift=0.2em},
yticklabel style={font=\scriptsize},
ytick={0.001, 0.01, 0.1, 1, 10, 100},
yticklabels={0.001, 0.01, 0.1, 1, 10, 100},
yminorticks=false,
ylabel={Success rate (\%)},
label style={font=\scriptsize},
ylabel style={at={(0.025,0.5)}},
legend style={font=\small, at={(0.5,1.25)}, anchor=north,legend columns=-1,/tikz/every even column/.append style={column sep=0.2cm}},
legend image code/.code={\draw [#1] (0cm,-0.1cm) rectangle (0.09cm,0.2cm);},
]

\addplot [index of colormap=4 of Accent, fill, ybar, bar width=0.09cm, xshift=-0.12cm] table [x expr=\coordindex, y={success_rate}]{\ours};
\addlegendentry{Ours}

\addplot [index of colormap=0 of Accent, fill, ybar, bar width=0.09cm] table [x expr=\coordindex, y={success_rate}]{\dreamer};
\addlegendentry{DreamerV3}

\addplot [index of colormap=6 of Paired, fill, ybar, bar width=0.09cm, xshift=0.12cm] table [x expr=\coordindex, y={success_rate}]{\lstm};
\addlegendentry{LSTM-SPCNN}

\end{axis}
\end{tikzpicture}
\caption{Individual success rates for all achievements.}
\label{fig:crafter_success_rate}
\end{figure}

\Cref{fig:crafter_success_rate} shows the individual success rates for all 22 achievements of our method and two successful baselines, DreamerV3 and LSTM-SPCNN.
Remarkably, our method outperforms the baselines in unlocking challenging achievements.
For instance, our method collects iron with a probability over 3\%, which is 20 times higher than DreamerV3.
This achievement is extremely challenging due to its scarcity on the map and the need for wood and stone tools.
Moreover, our method crafts iron tools with a probability of approximately 0.01\%, which is not achievable by either of the baselines.
Finally, our method even succeeds in collecting diamonds, the most challenging task, on individual runs.

\begin{table}[h]
\centering
\caption{Scores and rewards. MuZero + SPR\textsuperscript{\textdagger} denotes the results replicated from the original paper.}
\begin{tabular}{ccccc}
\toprule
\multicolumn{2}{c}{Method} & Parameters & Score (\%) & Reward  \\
\midrule
\multicolumn{2}{c}{Human Expert} & - & 50.5 $\pm$ 6.8 & 14.3 $\pm$ 2.3 \\
\midrule
\multirow{6}{*}{From scratch} & Ours & 9M & \textbf{21.79} $\pm$ \textbf{1.37} & \textbf{12.60} $\pm$ \textbf{0.31} \\ 
\cmidrule{2-5}
& PPO & 4M & 15.60 $\pm$ 1.66 & 10.32 $\pm$ 0.53 \\
& DreamerV3 & 201M & 14.77 $\pm$ 1.42 & 10.92 $\pm$ 0.53 \\ 
& LSTM-SPCNN & 135M & 11.67 $\pm$ 0.80 & \textcolor{white}{0}9.34 $\pm$ 0.23 \\
& MuZero + SPR\textsuperscript{\textdagger} & 54M & \textcolor{white}{0}4.4 $\pm$ 0.4 & \textcolor{white}{0}8.5 $\pm$ 0.1 \\
& SEA & 1.5M & \textcolor{white}{0}1.22 $\pm$ 0.13 & \textcolor{white}{0}0.63 $\pm$ 0.08 \\
\midrule
Pre-training & MuZero + SPR\textsuperscript{\textdagger} & 54M & 16.4 $\pm$ 1.5 & 12.7 $\pm$ 0.4 \\
\bottomrule
\end{tabular}
\label{tab:crafter_score}
\end{table}

\subsection{Model size analysis}

We compare the model sizes between our method and the baselines. As shown in \Cref{tab:crafter_score}, our method achieves better performance with fewer parameters.
In particular, our method only requires 4\% of the parameters used by DreamerV3.
Note that while our method has twice as many parameters as PPO, most of this increase is due to the networks used for the auxiliary learning, which are not utilized during inference.
Additionally, we test our method with a smaller model by reducing the channel size to [16, 32, 32] and the hidden dimension to 256, resulting in a total of 1M parameters. Notably, it still outperforms the baselines with a score of 17.07\%.

\subsection{Representation analysis of achievement distillation}

To validate whether our method induces the encoder to have better representations for predicting the next achievements, we conduct an analysis of the latent representations of the encoder using the same approach as described in \Cref{sec:motivation}. Our method achieves a classification accuracy of 73.6\%, which is a 28.7\%p increase compared to PPO. Furthermore, \Cref{fig:confidence} demonstrates that our method produces predictions with significantly higher confidence than PPO, with a median value of 0.752.

\subsection{Ablation studies}

\begin{wraptable}{r}{4.5cm}
\vspace{-2.5em}
\centering
\caption{Ablation studies.}
\begin{tabular}{cccc}
\toprule
I & C & M & Score (\%) \\
\midrule
\red{\ding{55}} & \red{\ding{55}} & \red{\ding{55}} & 15.60 $\pm$ 1.66 \\
\green{\ding{51}} & \red{\ding{55}} & \red{\ding{55}} & 19.02 $\pm$ 1.65 \\ 
\green{\ding{51}} & \green{\ding{51}} & \red{\ding{55}} & 20.36 $\pm$ 1.79 \\
\green{\ding{51}} & \green{\ding{51}} & \green{\ding{51}} & \textbf{21.79} $\pm$ \textbf{1.37} \\
\bottomrule
\end{tabular}
\label{tab:crafter_ablation_score}
\vspace{-2em}
\end{wraptable}

We conduct ablation studies to evaluate the individual contribution of our proposed method, intra-trajectory achievement prediction (I), cross-trajectory achievement matching (C), and memory (M).
\Cref{tab:crafter_ablation_score} shows that while intra-trajectory achievement prediction is the most significant contributor, cross-trajectory achievement matching and memory also play important roles in improving the performance of our method.

\subsection{Extension to value-based algorithms}

In our previous experiments, we use the on-policy policy gradient algorithm, PPO, as our backbone RL algorithm. To assess the adaptability of our method to other RL paradigms, we extend our experiments to the popular off-policy value-based algorithm, QR-DQN, introducing a slight modification \citep{dabney2018distributional}. Specifically, we employ Huber quantile regression to preserve the Q-network’s output distribution in alignment with the value function optimization in QR-DQN. We train the agent on Crafter for 1M environment steps and evaluate its performance. Notably, our method also proves effective for value-based algorithms, elevating the score from 4.14 to 8.07. The detailed experimental settings and results are provided in Appendix D.

\subsection{Application to other environments}

To evaluate the broad applicability of our method to diverse environments, we conduct experiments on two additional benchmarks featuring hierarchical achievements: Procgen Heist and MiniGrid \citep{cobbe2020leveraging,chevalier2023minigrid}. Heist is a procedurally generated environment whose goal is to steal a gem hidden behind a sequence of blue, green, and red locks. To open each lock, an agent must collect a key with the corresponding color. Heist introduces another challenge, given that the color of wall and background can vary between environments, whereas Crafter maintains fixed color patterns for its terrains. Additionally, we create a customized a door-key environment using MiniGrid to evaluate the effectiveness of our method on a deeper achievement graph. Notably, our method significantly improves the performance of PPO in Heist, increasing the score from 29.6 to 71.0. Our method also outperforms PPO in the MiniGrid environment by a substantial margin, elevating the score from 3.33 to 8.04. The detailed experimental settings and results can be found in Appendix E.

\section{Related work}

\paragraph{Discovering hierarchical achievements in RL}
One major approach to this problem is model-based algorithms.
DreamerV3 learns a world model that predicts future states and rewards and trains an agent using imagined trajectories generated by the model \citep{hafner2023mastering}.
While achieving superior performance on Crafter, it requires more than 200M parameters.
MuZero + SPR trains a model-based agent with a self-supervised task of predicting future states \citep{schrittwieser2021online,walker2023investigating}.
However, it relies on pre-training with 150M environment steps collected via RND to improve its performance on Crafter \cite{burda2018exploration}.

Another approach is hierarchical algorithms, while many of these methods are only tested on grid-world environments \citep{sohn2020meta,costales2022possibility}.
HAL introduces a classifier that predicts the next achievement to be done and uses it as a high-level planner \citep{costales2022possibility}.
However, it relies on prior information on achievements, such as what achievement has been completed, and does not scale to the high-dimensional Crafter.
SEA reconstructs the Crafter achievement graph using 200M offline data collected from a pre-trained IMPALA policy and employs a high-level planner on the graph \citep{zhou2023learning}.
However, it requires expert data to fully reconstruct the graph and has not been tested on a sample-efficient regime.

There are only a few studies that have explored model-free algorithms. LSTM-SPCNN uses an LSTM and a size-preserving CNN to improve the performance of PPO on Crafter \citep{hochreiter1997long,locatello2020object,stanic2022learning}.
However, it requires 135M parameters and the performance gain is modest compared to PPO with a CNN.

\paragraph{Representation learning in RL} 
There has been a large body of work on representation learning for improving sample efficiency on a single environment \citep{laskin2020curl,yarats2021reinforcement,schwarzer2021dataefficient}, or generalization on procedurally generated environments \citep{mazoure2020deep,mazoure2022crosstrajectory,moon2022rethinking}. However, representation learning for discovering hierarchical achievements has little been explored.
A recent study has evaluated the performance of the previously developed representation learning technique SPR on Crafter, but the results are not promising \citep{schwarzer2021dataefficient,walker2023investigating}.

While widely used in other domains, optimal transport has recently garnered attention in RL \cite{dadashi2021primal,fickinger2022crossdomain,luo2023optimal}. However, the majority of the studies focus on imitation learning, where optimal transport is employed to match the behavior of an agent with offline expert data. In this work, we utilize optimal transport to obtain generalizable representations for achievements in the online setting.

\section{Conclusion}

In this work, we introduce a novel self-supervised method for discovering hierarchical achievements, named \emph{achievement distillation}.
This method distills relevant information about achievements from episodes collected during policy updates into the encoder and can be seamlessly integrated with a popular model-free algorithm PPO.
We show that our proposed method is capable of discovering the hierarchical structure of achievements without any explicit component for long-term planning, achieving state-of-the-art performance on the Crafter benchmark using fewer parameters and data.

While we utilize only minimal information about hierarchical achievements (\ie, the agent receives a reward when a new achievement is unlocked), one limitation of our work is that we have not evaluated the transferability of our method to an unsupervised agent without any reward.
A promising future direction would be developing a representation learning method that can distinguish achievements in a fully unsupervised manner and combining it with curiosity-driven exploration techniques \citep{sekar2020planning,guo2022byol}.

\section*{Acknowledgements}
This work was partly supported by Institute of Information \& Communications Technology Planning \& Evaluation (IITP) grant funded by the Korea government (MSIT) (No. 2020-0-00882, (SW STAR LAB) Development of deployable learning intelligence via self-sustainable and trustworthy machine learning, 80\%, and No. 2022-0-00480, Development of Training and Inference Methods for Goal-Oriented Artificial Intelligence Agents, 20\%). This research was supported by a grant from KRAFTON AI. This material is based upon work supported by the Air Force Office of Scientific Research under award number FA2386-23-1-4047. Hyun Oh Song is the corresponding author.

\newpage

\bibliography{main}
\bibliographystyle{plainnat}

\newpage

\appendix

\section{Representation analysis}
\label{sec:rep}

To analyze the latent representations learned by PPO and our method, we construct a dataset using an expert policy. Specifically, we initially train the expert policy using our method with 1M environment steps. Note that this policy is trained with a different seed from the policies used for evaluation. Next, we collect a batch of episodes using the expert policy, resulting in a dataset containing 215,578 states. From this dataset, we subsample 50,000 states for the training set and 10,000 states for the test set.

For each method, we acquire the latent representations from the encoder for the training set and freeze them. We then train a linear classifier using these representations to predict the very next achievements. Each achievement is labeled from 0 to 21, representing the different possible achievements in Crafter. We optimize the classifier for 500 epochs using the Adam optimizer with a learning rate of 1e-3 \citep{kingma2014adam}. Finally, we measure the classification accuracy and the prediction probability (\ie, confidence) for the ground-truth label on the test set.

\section{Examples of cross-trajectory achievement matching}

To demonstrate the effectiveness of cross-trajectory achievement matching, we provide an example of matching results for our method and PPO.
We first collect two different episodes using an expert policy, following the same procedure outlined in \Cref{sec:rep}.
Subsequently, we acquire the representations of the achievement sequences for each method. Ultimately, we perform the matching process between the sequences of achievement representations for each method.

\begin{figure}[ht!]
\centering
\begin{subfigure}[b]{0.49\textwidth}
\includegraphics[width=\textwidth]{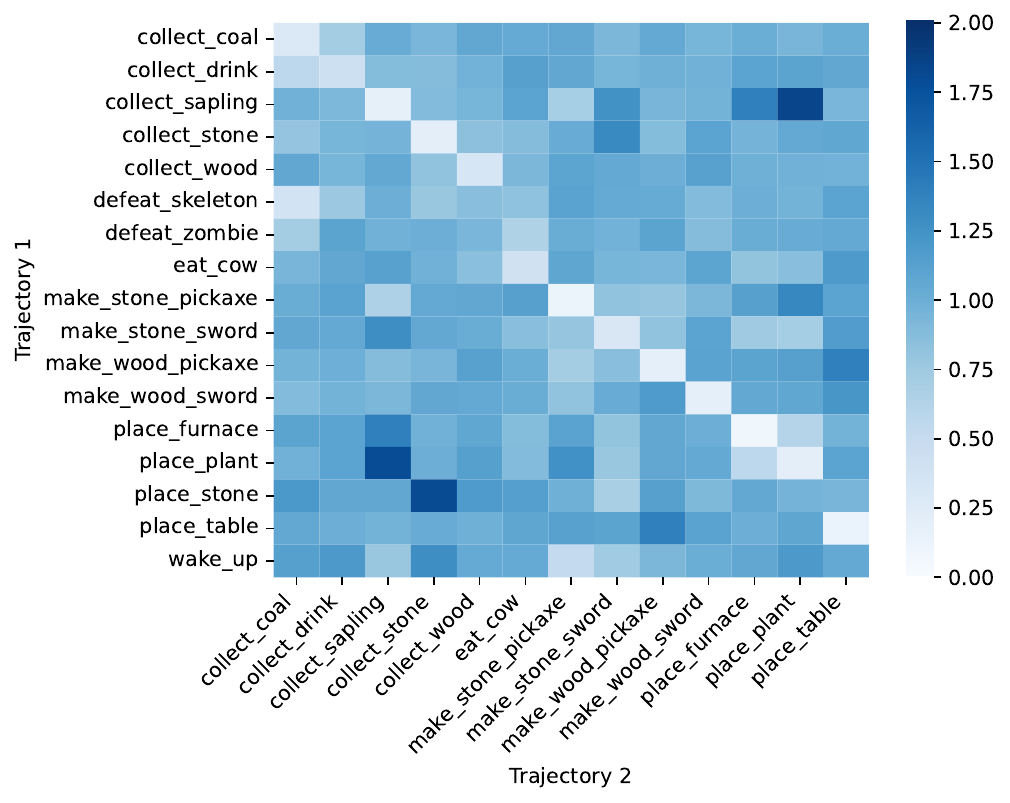}
\caption{Ours}
\label{fig:cosine_distance_ours}
\end{subfigure}
\hfill
\begin{subfigure}[b]{0.49\textwidth}
\includegraphics[width=\textwidth]{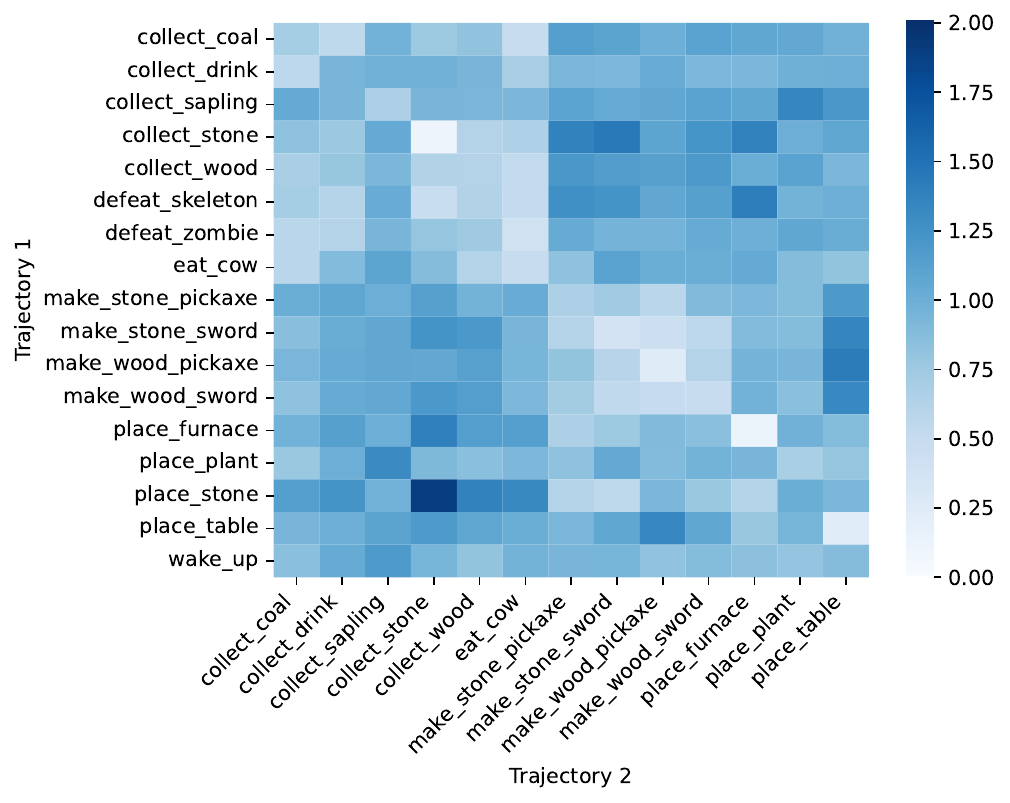}
\caption{PPO}
\label{fig:cosine_distance_ppo}
\end{subfigure}
\caption{Cosine distance between achievement representations.}
\label{fig:cosine_distance}
\end{figure}

\Cref{fig:cosine_distance} visualizes the cosine distance between the achievement representations from the two episodes.
Remarkably, our method exhibits a lower cosine distance between the same achievements compared to PPO.
This highlights the effectiveness of cross-trajectory achievement matching in facilitating the learning of generalizable representations for achievements across different episodes.

\begin{figure}[ht!]
\centering
\begin{subfigure}[b]{0.49\textwidth}
\includegraphics[width=\textwidth]{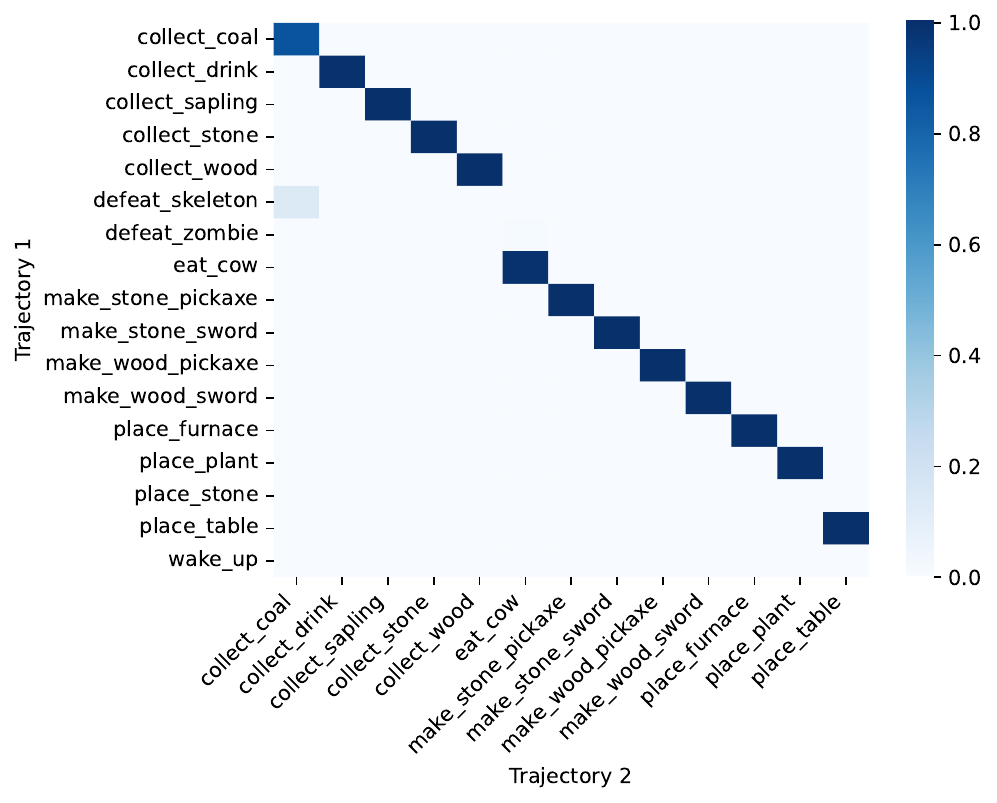}
\caption{Ours}
\label{fig:soft_matching_ours}
\end{subfigure}
\hfill
\begin{subfigure}[b]{0.49\textwidth}
\includegraphics[width=\textwidth]{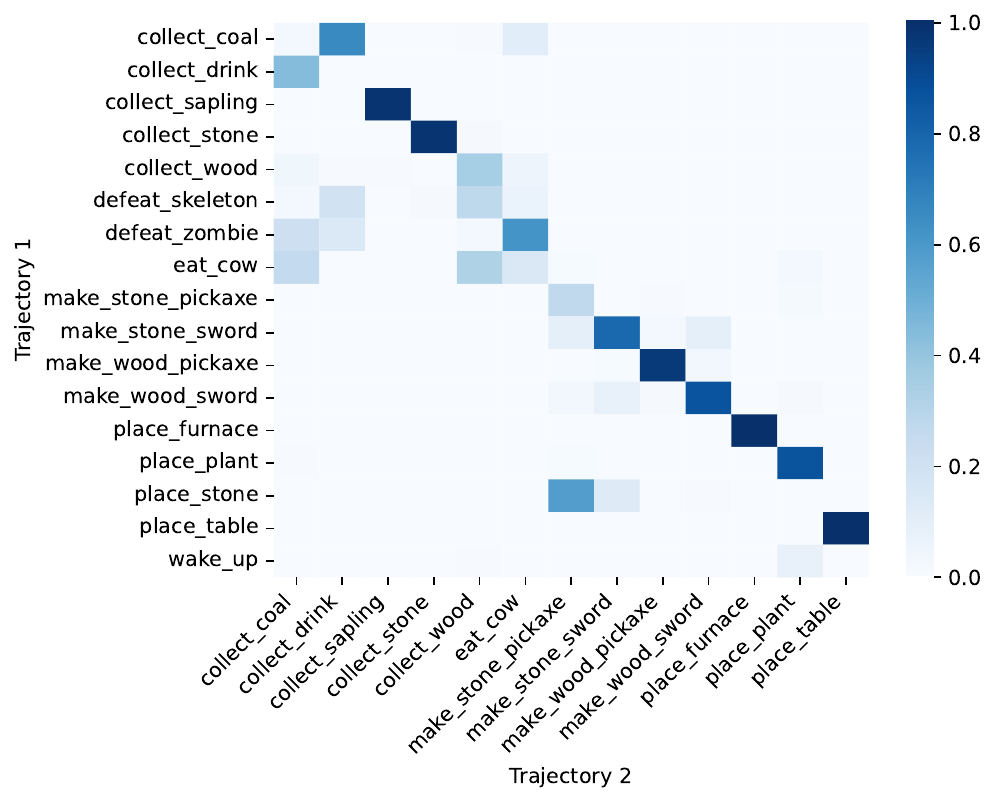}
\caption{PPO}
\label{fig:soft_matching_ppo}
\end{subfigure}
\caption{Soft matching via partial optimal transport.}
\label{fig:soft_matching}
\end{figure}

\begin{figure}[ht!]
\centering
\begin{subfigure}[b]{0.49\textwidth}
\includegraphics[width=\textwidth]{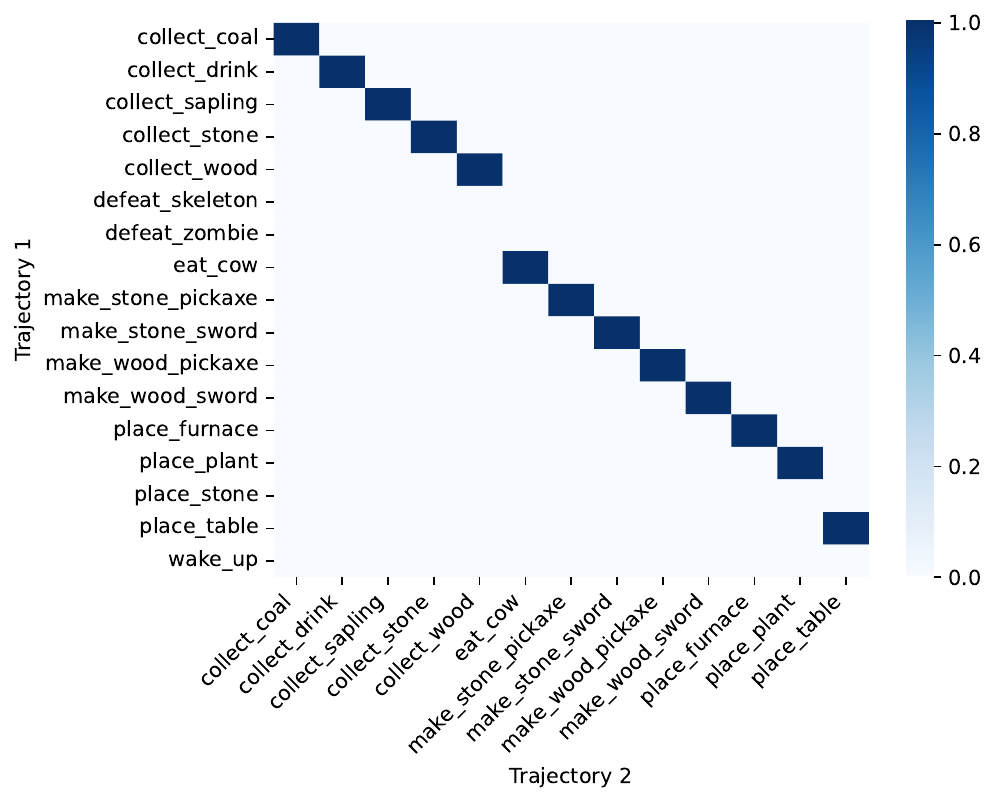}
\caption{Ours}
\label{fig:hard_matching_ours}
\end{subfigure}
\hfill
\begin{subfigure}[b]{0.49\textwidth}
\includegraphics[width=\textwidth]{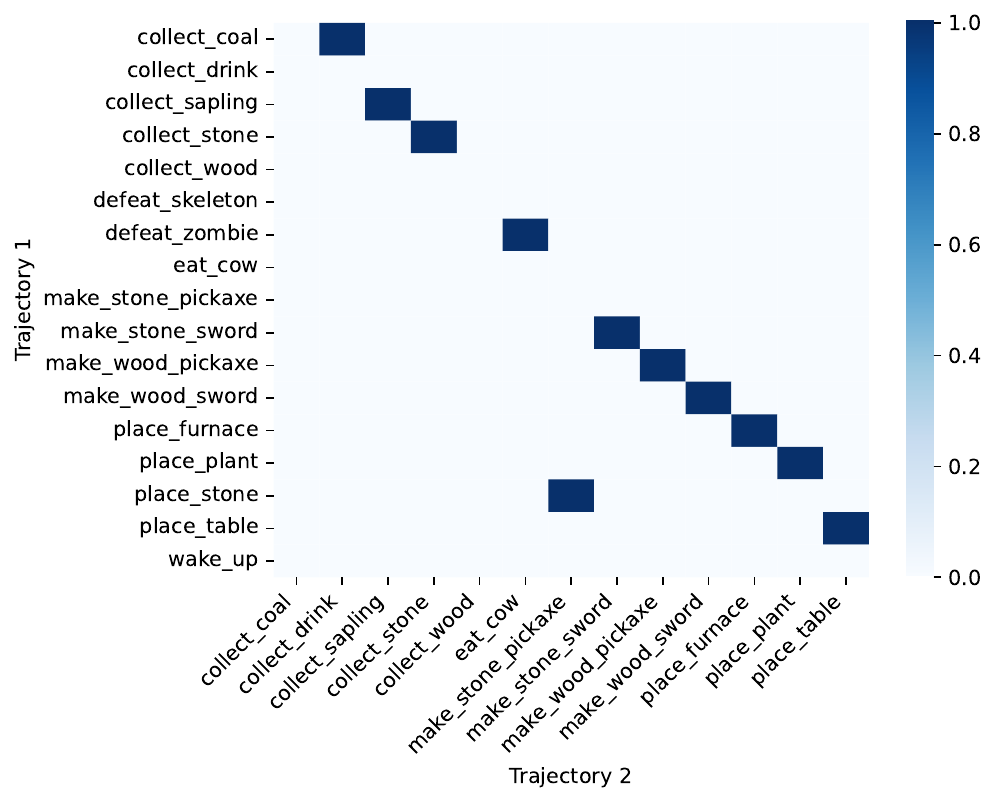}
\caption{PPO}
\label{fig:hard_matching_ppo}
\end{subfigure}
\caption{Hard matching via thresholding.}
\label{fig:hard_matching}
\end{figure}

\Cref{fig:soft_matching} illustrates the soft matching results computed using partial optimal transport \citep{benamou2015iterative}.
Notably, the matching result of PPO contains inaccurate or unconfident matchings, while our method does not suffer from this issue.
Consequently, the hard matching result of PPO exhibits inaccurate matchings, as depicted in \Cref{fig:hard_matching_ppo}.
For instance, ``Defeat zombie'' in the first episode is matched with ``Eat cow' in the second episode and ``Collect wood'' is not matched at all.
In contrast, \Cref{fig:hard_matching_ours} shows that our method successfully matches identical achievements between episodes.
Furthermore, our method avoids matching achievements in one episode that do not exist in the other episode.

\clearpage
\newpage

\begin{figure}[ht!]
\centering
\begin{subfigure}[b]{0.49\textwidth}
\includegraphics[width=\textwidth]{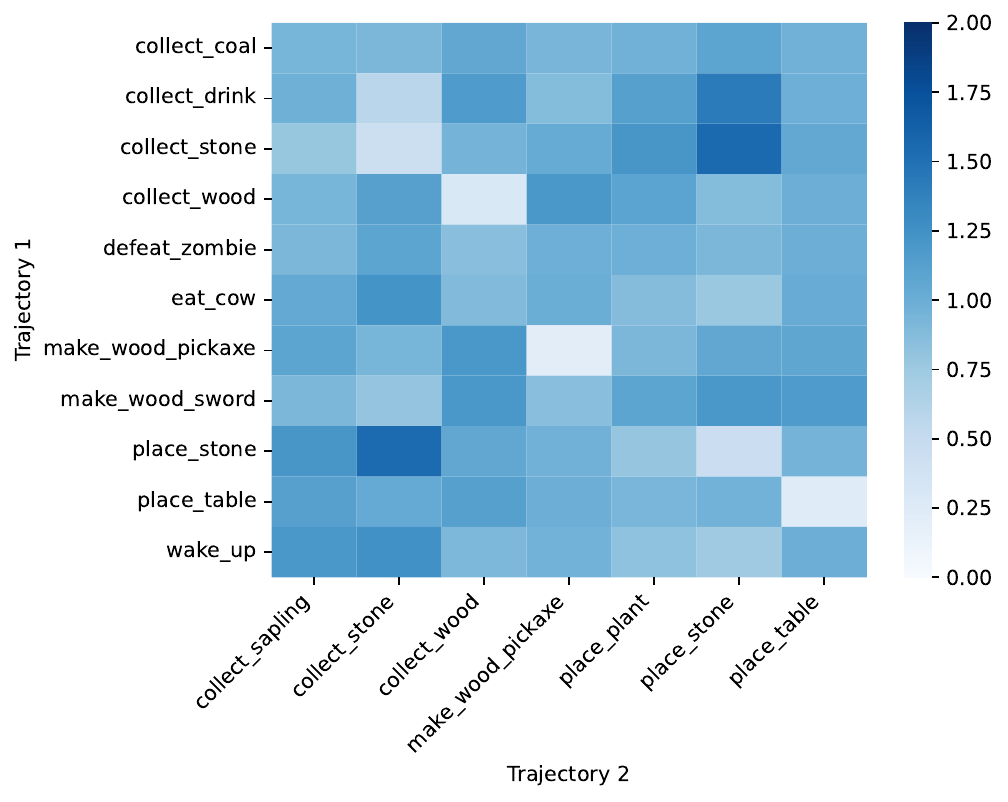}
\caption{Cosine distance}
\label{fig:distance_100}
\end{subfigure}
\hfill
\begin{subfigure}[b]{0.49\textwidth}
\includegraphics[width=\textwidth]{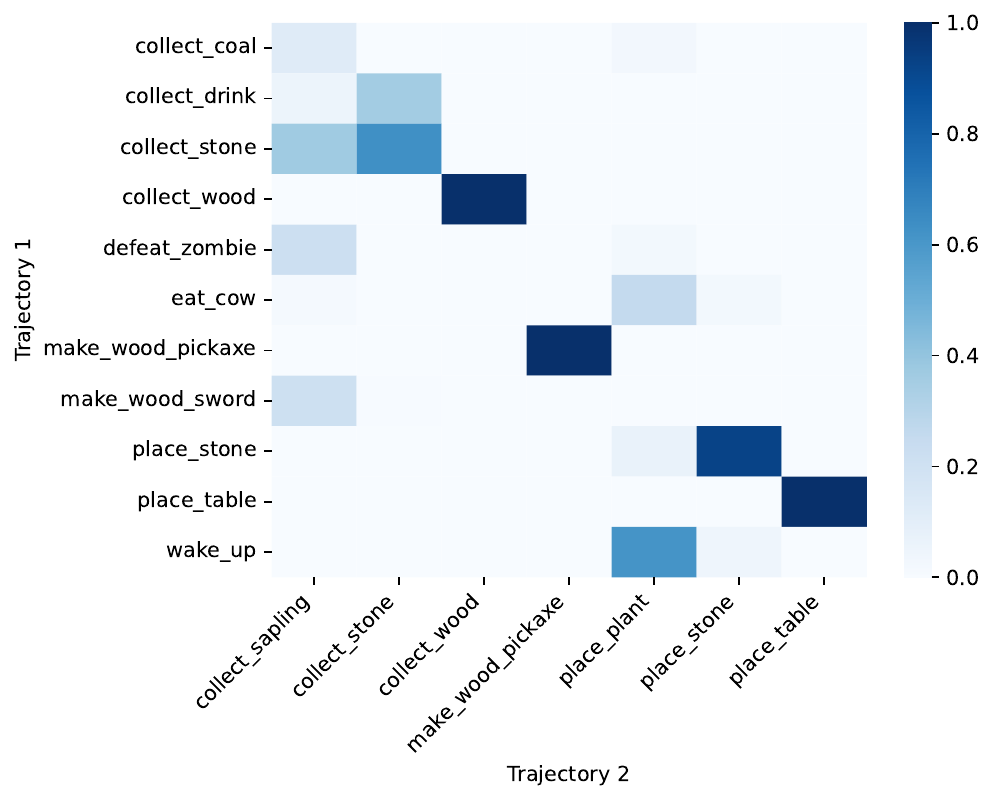}
\caption{Soft matching via partial optimal transport (ours)}
\label{fig:softmatching_100}
\end{subfigure}

\bigskip

\begin{subfigure}[b]{0.49\textwidth}
\includegraphics[width=\textwidth]{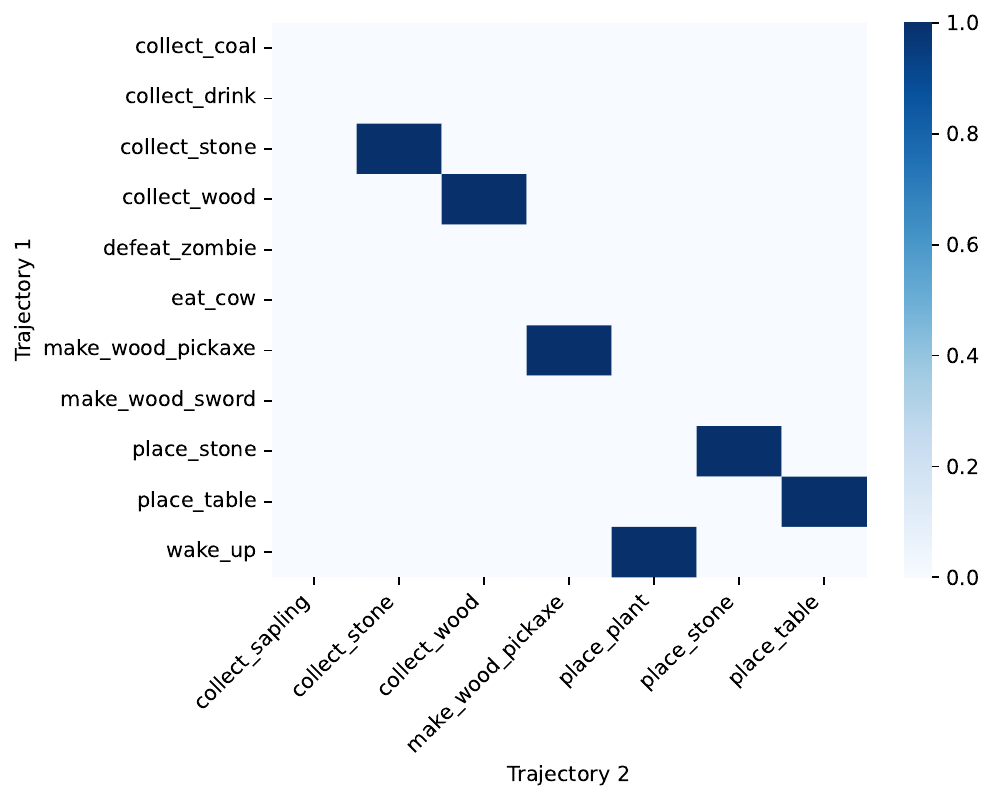}
\caption{Hard matching via thresholding (ours)}
\label{fig:hardmatching_100}
\end{subfigure}
\hfill
\begin{subfigure}[b]{0.49\textwidth}
\includegraphics[width=\textwidth]{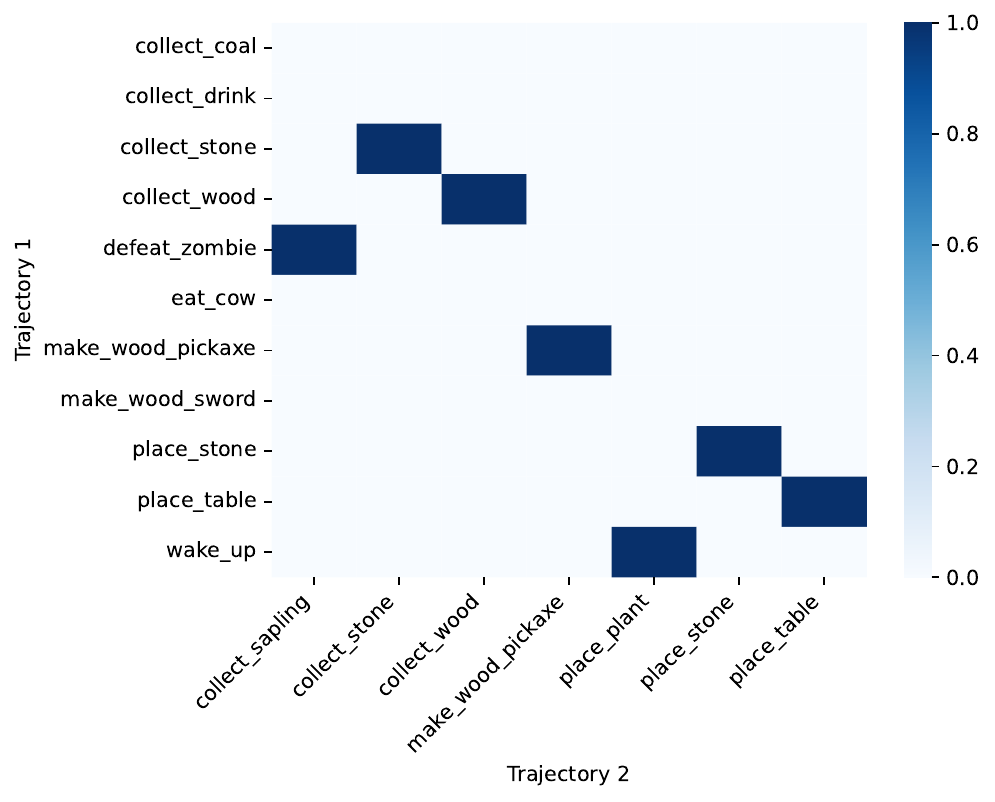}
\caption{Hard matching via Hungarian algorithm}
\label{fig:hardmatching_100_bipar}
\end{subfigure}
\caption{Matching results of our matching algorithm and the Hungarian algorithm.}
\label{fig:bipartite}
\end{figure}

We further explore the effectiveness of our matching algorithm that utilizes partial optimal transport followed by thresholding during the early stage of training. Specifically, we first collect two different episodes using a policy trained with our method for up to 100 epochs. Subsequently, we employ our matching algorithm to calculate the hard matching between the achievement sequences extracted from these episodes and compare it with the Hungarian algorithm, which is commonly used in bipartite graph matching \cite{munkres1957algorithms}.

\Cref{fig:bipartite} provides an example of the matching results obtained using our matching algorithm and the Hungarian algorithm. In the case of the Hungarian algorithm, ``Defeat zombie'' in the first episode is incorrectly matched with ``Collect sapling'' in the second episode, as shown in \Cref{fig:hardmatching_100_bipar}. In contrast, our matching algorithm effectively avoids matching ``Defeat zombie'' in the first episode, as depicted in \Cref{fig:hardmatching_100}.

\clearpage
\newpage

\section{Experimental settings}

\subsection{Computational resources}
All experiments are conducted on an internal cluster, with each node consisting of two AMD EPYC 7402 CPUs, 500GB of RAM, and eight NVIDIA RTX 3090 GPUs. We utilize PyTorch as our primary deep learning framework \citep{paszke2019pytorch}.

\subsection{Implementation details and hyperparameters}
\paragraph{PPO} Our implementation of PPO is based on the official code repository (\url{https://github.com/openai/Video-Pre-Training}) provided by \citet{baker2022video}.
Each input image has dimensions of $64 \times 64 \times 3$.
The image is first processed with a ResNet encoder as proposed in IMPALA, which consists of three stacks with channel sizes of $[64, 64, 128]$ \citep{espeholt2018impala}.
Each stack is composed of a $3 \times 3$ convolutional layer with a stride of 1, a $3 \times 3$ max pooling layer with a stride of 2, and two ResNet blocks as introduced in \citet{he2016deep}.
Subsequently, the output of the encoder is flattened into a vector of size 8192 and passed through two consecutive dense layers with output sizes of 256 and 1024, respectively.
This resulting vector serves as the latent representation.
Finally, the latent representation is fed into two independent dense layers, the policy and value heads.
The policy head has an output size of 17 and generates a categorical distribution over the action space.
The value head has an output size of 1 and produces a scalar value representing the value function.
All weights are initialized using fan-in initialization, except for the policy and value heads \citep{lecun2002efficient}.
The weights of the policy and value heads are initialized using orthogonal initialization with a gain of $0.01$ and $0.1$, respectively \citep{saxe2013exact}.
All biases are initialized to zero.
We use ReLU as the activation function \citep{nair2010rectified}. All network parameters are optimized using the Adam optimizer \citep{kingma2014adam}.

For PPO training, we slightly modify the hyperparameter setting from \citet{stanic2022learning}.
Specifically, we decrease the number of mini-batches per epoch from 32 to 8 and the number of epochs per rollout from 4 to 3. These modified settings are commonly used in \citet{cobbe2021phasic,moon2022rethinking}.
Instead of employing the reward normalization technique proposed in the original PPO paper, we normalize the value function target with the mean and standard deviation estimated through an exponentially weighted moving average (EWMA), following the practice in \citet{baker2022video}. We set the decay rate of the EWMA to 0.99.
We provide the default values for the hyperparameters in \Cref{tab:ppo_hp}.

\begin{table}[h]
\centering
\caption{PPO hyperparameters.}
\begin{tabular}{cc}
\toprule
Hyperparameter & Value \\
\midrule
Discount factor & 0.95 \\
GAE smoothing parameter & 0.65 \\
\# timesteps per rollout & 4096 \\
\# epochs per rollout & 3 \\
\# mini-batches per epoch & 8 \\
Entropy bonus  & 0.01 \\
PPO clip range & 0.2 \\
Reward normalization & No \\
EWMA decay rate & 0.99 \\
Learning rate & 3e-4 \\
Max grad norm & 0.5 \\
Value function coefficient & 0.5 \\
\bottomrule
\end{tabular}
\label{tab:ppo_hp}
\end{table}

\paragraph{DreamerV3}

To reproduce the results, we utilize the official code repository (\url{https://github.com/danijar/dreamerv3}) provided by \citet{hafner2023mastering}. We use the recommended hyperparameter setting from the original paper. Specifically, we set the model size to XL and the training ratio, which is the number of imagined steps per environment step, to 512. For more detailed information about the hyperparameter, please refer to \Cref{tab:dreamer_hp}.

\begin{table}[h]
\centering
\caption{DreamerV3 hyperparameters.}
\begin{tabular}{cc}
\toprule
Hyperparameter & Value \\
\midrule
GRU recurrent units & 4096 \\
CNN multiplier & 96 \\
Dense hidden units & 1024 \\
MLP layers & 5 \\
Training ratio & 512 \\
\bottomrule
\end{tabular}
\label{tab:dreamer_hp}
\end{table}

\paragraph{LSTM-SPCNN}

To reproduce the results, we utilize the official code repository (\url{https://github.com/astanic/crafter-ood}) provided by \citet{stanic2022learning}.
Regarding the network architecture, a size-preserving CNN is used as the image encoder \citep{locatello2020object}.
This encoder consists of four convolutional layers, each with a kernel size of $5 \times 5$, a channel size of 64, and a stride of 1, and does not have any pooling layers.
The encoder output is flattened into a vector of size 262144 and fed into a dense layer with an output size of 512, yielding the latent representation.
For the policy network, the latent representation is first passed through an LSTM layer with a hidden size of 256 and then fed into a dense layer with an output size of 17, producing a categorical distribution over the actions \citep{hochreiter1997long}.
For the value network, the latent representation is passed through two consecutive dense layers with output sizes of 256 and 1, respectively, to generate a scalar value representing the value function.
For PPO training, we adopt the best hyperparameter setting from the original paper. Please refer to \Cref{tab:lstm_spcnn_hp} for the default values for the hyperparameters.

\begin{table}[h]
\centering
\caption{LSTM-SPCNN hyperparameters.}
\begin{tabular}{cc}
\toprule
Hyperparameter & Value \\
\midrule
Discount factor & 0.95 \\
GAE smoothing parameter & 0.65 \\
\# timesteps per rollout & 4096 \\
\# epochs per rollout & 4 \\
\# minibatches per epoch & 32 \\
Entropy bonus  & 0.0 \\
PPO clip range & 0.2 \\
Reward normalization & No \\
Learning rate & 3e-4 \\
Max grad norm & 0.5 \\
Value function coefficient & 0.5 \\
\bottomrule
\end{tabular}
\label{tab:lstm_spcnn_hp}
\end{table}

\paragraph{MuZero + SPR}

We report the results replicated from the original paper since the official code has not been released yet \citep{walker2023investigating}. It is worth noting that the paper uses an increased resolution for the input images, from $64 \times 64$ to $96 \times 96$, which can potentially result in improved performance compared to the original settings.

\paragraph{SEA}

To evaluate its performance, we utilize the official code repository (\url{https://github.com/pairlab/iclr-23-sea}) provided by \citet{zhou2023learning}. 
It is important to mention that the paper uses a modified version of Crafter, where the health mechanism and associated rewards are removed.
This modification allows the agent to be immortal and explore the world map without any constraints, making the environment much easier than the original.
Furthermore, the paper trains an agent using a substantial number of 500M environment steps.
The paper uses 200M environment steps to train the IMPALA policy for offline data collection, followed by additional 300M environment steps to train the sub-policies for exploration.

To ensure a fair comparison with other methods, we reproduce the results by training an agent on the original Crafter benchmark using 1M environment steps. 
Specifically, we use 400K environment steps for training the IMPALA policy and 600K environment steps for training the sub-policies, maintaining the same ratio as the original setting.
Regarding the network architecture, we keep it unchanged from the original implementation. However, it is worth noting that the original implementation employs a higher resolution of $84 \times 84$ for the input images, which can potentially improve the performance. We adopt the best hyperparameter setting for training from the original paper. Please refer to \Cref{tab:sea_hp} for the default values for the hyperparameters.

\begin{figure}[t]
\centering
\begin{subfigure}[b]{0.33\textwidth}
\includegraphics[width=\textwidth]{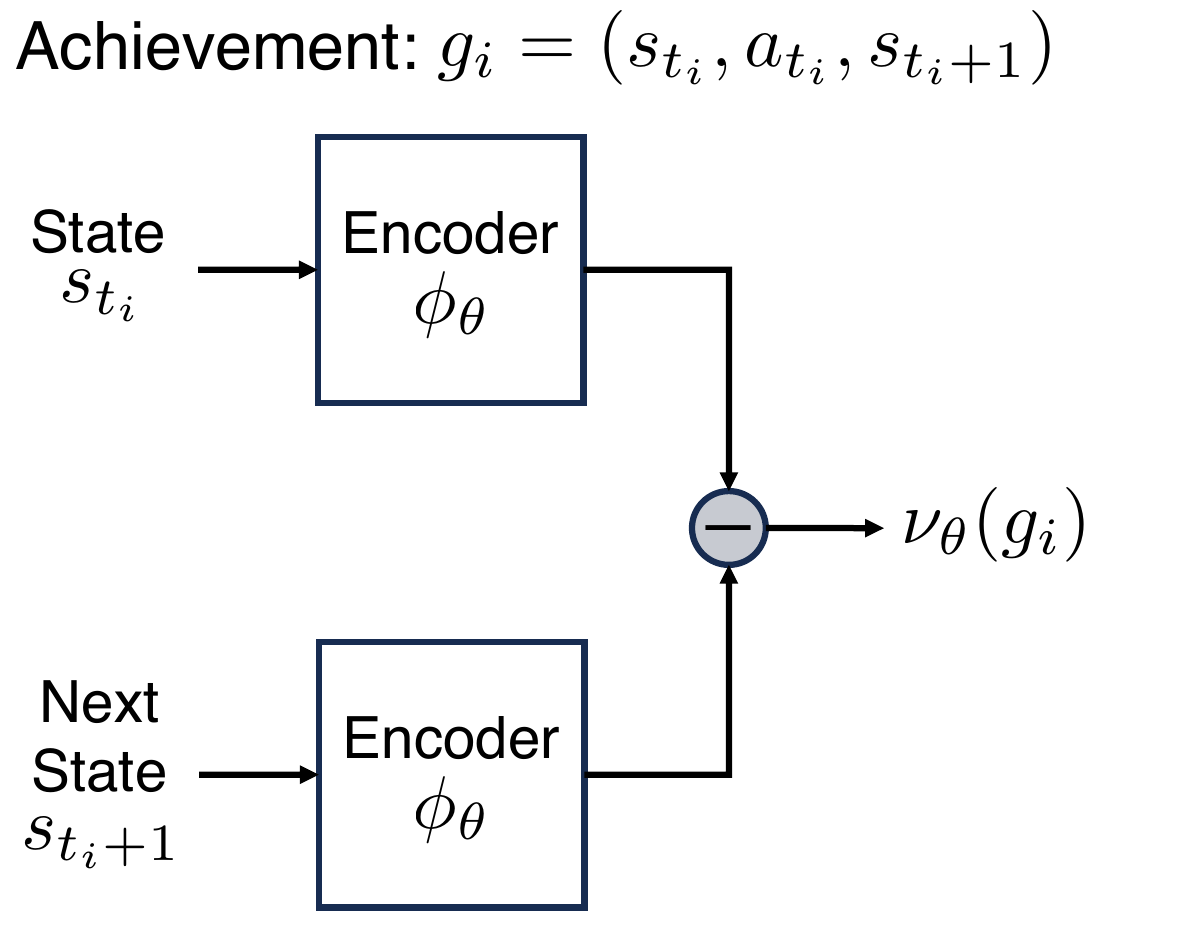}
\caption{Achievement representation}
\label{fig:network_g}
\end{subfigure}
\hfill
\begin{subfigure}[b]{0.66\textwidth}
\includegraphics[width=\textwidth]{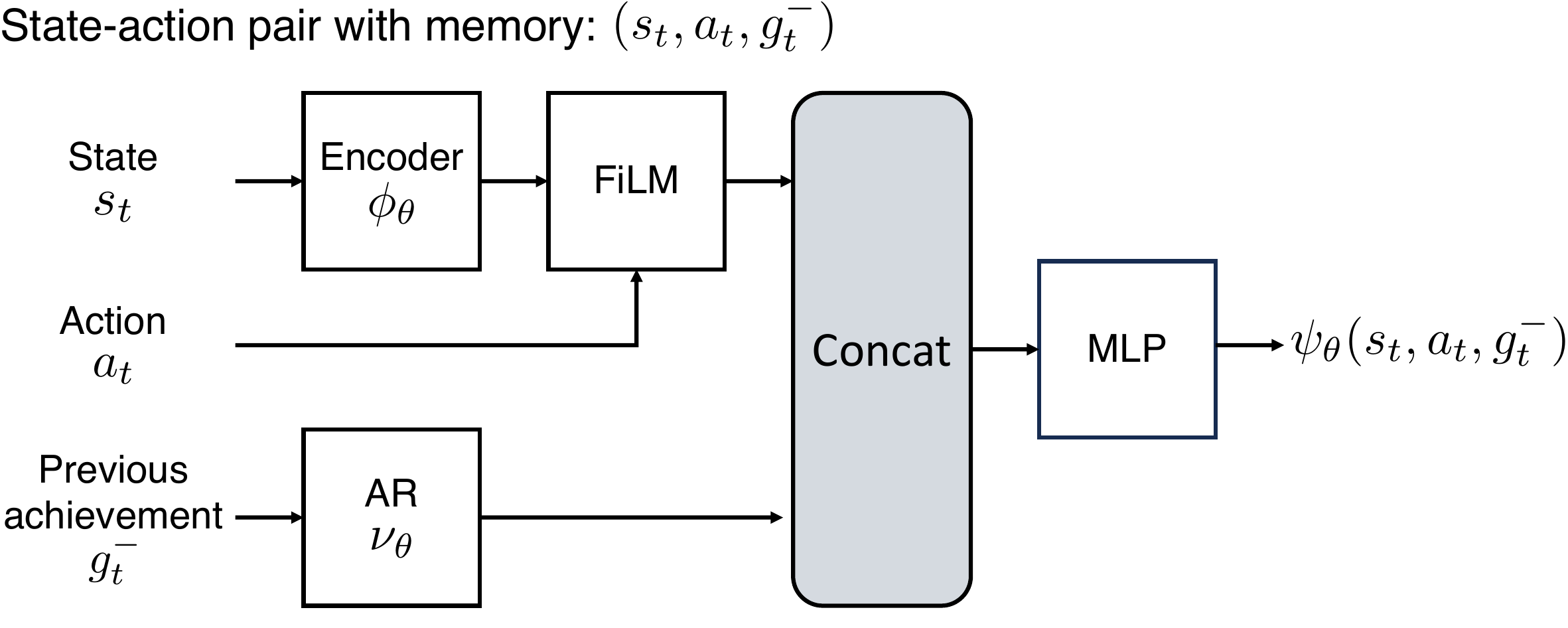}
\caption{State-action pair representation}
\label{fig:network_sa}
\end{subfigure}
\caption{Network architectures for the representations of (a) achievements and (b) state-action pairs.}
\label{fig:network}
\end{figure}

\begin{table}[h]
\centering
\caption{SEA hyperparameters.}
\begin{tabular}{cc}
\toprule
Hyperparameter & Value \\
\midrule
Network architecture & CNN + LSTM \\
Hidden size & 256 \\
Learning rate & 2e-4 \\
Batch size & 32 \\
Unroll length & 80 \\
Gradient clipping & 40 \\
RMSProp $\alpha$ & 0.99 \\
RMSProp momentum & 0 \\
RMSProp $\epsilon$ & 0.01 \\
Discount factor & 0.99 \\
Reward normalization & Yes \\
\# timesteps for IMPALA policy training & 400K \\
\# timesteps for achievement classifier training & 100K \\
\# timesteps for sub-policy training & 600K \\
\bottomrule
\end{tabular}
\label{tab:sea_hp}
\end{table}

\paragraph{Achievement distillation} 

Our method is built upon the PPO implementation, utilizing the same network architecture and hyperparameters for PPO training.
For achievement distillation, we compute the representation of an achievement $g_i = (s_{t_i}, a_{t_i}, s_{t_i + 1})$ by computing the difference between the latent state representations from the encoder:
\begin{align*}
\nu_\theta(g_i) = \phi_\theta(s_{t_i + 1}) - \phi_\theta(s_{t_i}),
\end{align*}
which is then normalized.
This process is illustrated in \Cref{fig:network_g}.
We also compute the representation of a state-action pair with the previous achievement as memory $(s_t, a_t, g_t^-)$ as
\begin{align*}
\psi_\theta(s_t, a_t, g_t^-) = \mathrm{MLP}_\theta(\mathrm{Concat}(\mathrm{FiLM}_\theta(\phi_\theta(s_t), a_t), \nu_\theta(g_t^-))),
\end{align*}
as described in \Cref{fig:network_sa}.
Specifically, the latent state representation from the encoder is combined with the action using a FiLM Layer:
\begin{align*}
\mathrm{FiLM}_\theta(\phi_\theta(s_t), a_t) = (1 + \eta_\theta(a_t)) \phi_\theta(s_t) + \delta_\theta(a_t),
\end{align*}
where $\eta_\theta$ and $\delta_\theta$ are two-layer MLPs, each with a hidden size of 1024 \citep{perez2018film}.
The resulting vector is then concatenated with the achievement representation and passed through a two-layer MLP with a hidden size of 1024, followed by normalization. 

When optimizing the contrastive objectives for achievement prediction and achievement matching, we jointly optimize the policy and value regularizer with the policy regularizer coefficient $\beta_\pi = 1.0$ and the value regularizer coefficient $\beta_V = 1.0$.
We compute a soft-matching between two achievement sequences using partial optimal transport with the entropic regularizer coefficient $\alpha=0.05$.
Note that these values are set without any hyperparameter search.

For the policy and auxiliary phases, we search for the number of policy phases per auxiliary phase within the range of $\{4, 8, 16\}$ and find $N_\pi = 8$ is optimal.
Similarly, we sweep over different values for the number of epochs per auxiliary phase, considering values of $\{1, 3, 6 \}$, and determine $E_\mathrm{aux} = 6$ as the optimal choice.
We provide the default values for the hyperparameter in \Cref{tab:ours_hp}.

\begin{table}[h]
\centering
\caption{Achievement distillation hyperparameters.}
\begin{tabular}{cc}
\toprule
Hyperparameter & Value \\
\midrule
Policy regularizer coefficient ($\beta_\pi$) & 1.0 \\
Value regularizer coefficient ($\beta_V$) & 1.0 \\
Entropic regularizer coefficient ($\alpha$) & 0.05 \\
\# policy phases per auxiliary phase ($N_\pi$)& 8 \\
\# epochs per auxiliary phase ($E_\mathrm{aux}$) & 6 \\
\bottomrule
\end{tabular}
\label{tab:ours_hp}
\end{table}

\subsection{Environment details}

\paragraph{Observation space}

An agent receives an image observation of dimensions $64 \times 64 \times 3$. This image contains a local, agent-centric view of the world map and the inventory state of the agent, such as the quantities of resources and tools and the levels of health, food, water, and energy.

\paragraph{Action space}

Crafter features a discrete 17-dimensional action space. 
The complete list of possible actions is provided in \Cref{tab:crafter_action_space}. The ``Do'' action encompasses activities including resource collection, consumption of food and water, and combat against enemies.

\begin{table}[h]
\centering
\caption{Crafter action space.}
\begin{tabular}{ccc}
\toprule
Index & Name \\
\midrule
0 & Noop \\
1 & Move left \\
2 & Move right  \\
3 & Move up \\
4 & Move down \\
5 & Do \\
6 & Sleep \\
7 & Place stone \\
8 & Place table \\
9 & Place furnace \\
10 & Place plant \\
11 & Make wood pickaxe \\
12 & Make stone pickaxe \\
13 & Make iron pickaxe \\
14 & Make wood sword \\
15 & Make stone sword \\
16 & Make iron sword \\
\bottomrule
\end{tabular}
\label{tab:crafter_action_space}
\end{table}

\paragraph{Achievements}

Crafter consists of 22 achievements that the agent can unlock by satisfying specific requirements. The complete list of the achievements and their corresponding requirements is presented in \Cref{tab:crafter_achievements}.

\begin{table}[h]
\centering
\caption{Crafter achievements and their requirements.}
\begin{tabular}{ccc}
\toprule
Name & Requirements \\
\midrule
Collect coal & Nearby coal; wood pickaxe \\
Collect diamond & Nearby diamond; iron pickaxe \\
Collect drink & Nearby water \\
Collect iron & Nearby iron; stone pickaxe  \\
Collect sapling & None \\
Collect stone & Nearby stone; wood pickaxe \\
Collect wood & Nearby wood \\
Defeat skeleton & None \\
Defeat zombie & None \\
Eat cow & None \\
Eat plant & Nearby plant \\
Make iron pickaxe & Nearby table and furnace; wood, coal, and iron \\
Make iron sword & Nearby table and furnace; wood, coal, and iron \\
Make stone pickaxe & Nearby table; wood and stone \\
Make stone sword & Nearby table; wood and stone \\
Make wood pickaxe & Nearby table; wood \\
Make wood sword & Nearby table; wood \\
Place furnace & Stone \\
Place plant & Sapling \\
Place stone & Stone \\
Place table & Wood \\
Wake up & None \\
\bottomrule
\end{tabular}
\label{tab:crafter_achievements}
\end{table}

\newpage

\section{Extension to value-based algorithms}

\begin{figure}[ht]
\centering
\begin{tikzpicture}
\begin{groupplot}[
group style = {group size = 2 by 1, horizontal sep=1.5cm},
width=5.5cm,
height=5cm,
grid=major,
xmin=0,
xmax=1e6,
scaled x ticks = false,
tick pos = left,
xtick={0, 2e5, 4e5, 6e5, 8e5, 10e5},
xticklabels={0, 0.2, 0.4, 0.6, 0.8, 1.0},
xlabel={Environment steps (M)},
ylabel near ticks,
xlabel style={font=\small},
ylabel style={font=\small},
]
\nextgroupplot[
ylabel=Score (\%),
no markers,
legend style={font=\small,legend columns=1},
legend to name=exp,
ytick={0, 2, 4, 6, 8, 10},
ymin=0.0,
ymax=10,
]

\addlegendimage{index of colormap=4 of Accent,line width=0.4mm}
\addlegendentry{Ours};

\addlegendimage{index of colormap=4 of Paired,line width=0.4mm}
\addlegendentry{QR-DQN};

\addplot[index of colormap=4 of Accent,line width=0.4mm] table [x=timesteps, y=scores_mean, col sep=comma]{qrdqn/data/qrdqn_ours_v2_score.csv};
\addplot[index of colormap=4 of Accent,name path=plus,draw=none] table [x=timesteps, y=scores_mean_plus_std, col sep=comma]{qrdqn/data/qrdqn_ours_v2_score.csv};
\addplot[index of colormap=4 of Accent,name path=minus,draw=none] table [x=timesteps, y=scores_mean_minus_std, col sep=comma]{qrdqn/data/qrdqn_ours_v2_score.csv};
\addplot[index of colormap=4 of Accent,fill opacity=0.15] fill between[of=plus and minus];

\addplot[index of colormap=4 of Paired,line width=0.4mm] table [x=timesteps, y=scores_mean, col sep=comma]{qrdqn/data/qrdqn_score.csv};
\addplot[index of colormap=4 of Paired,name path=plus,draw=none] table [x=timesteps, y=scores_mean_plus_std, col sep=comma]{qrdqn/data/qrdqn_score.csv};
\addplot[index of colormap=4 of Paired,name path=minus,draw=none] table [x=timesteps, y=scores_mean_minus_std, col sep=comma]{qrdqn/data/qrdqn_score.csv};
\addplot[index of colormap=4 of Paired,fill opacity=0.15] fill between[of=plus and minus];

\nextgroupplot[
ylabel=Reward,
no markers,
ymin=0,
ymax=10,
]

\addplot[index of colormap=4 of Accent,line width=0.4mm] table [x=timesteps, y=rewards_mean, col sep=comma]{qrdqn/data/qrdqn_ours_v2_score.csv};
\addplot[index of colormap=4 of Accent,name path=plus,draw=none] table [x=timesteps, y=rewards_mean_plus_std, col sep=comma]{qrdqn/data/qrdqn_ours_v2_score.csv};
\addplot[index of colormap=4 of Accent,name path=minus,draw=none] table [x=timesteps, y=rewards_mean_minus_std, col sep=comma]{qrdqn/data/qrdqn_ours_v2_score.csv};
\addplot[index of colormap=4 of Accent,fill opacity=0.15] fill between[of=plus and minus];

\addplot[index of colormap=4 of Paired,line width=0.4mm] table [x=timesteps, y=rewards_mean, col sep=comma]{qrdqn/data/qrdqn_score.csv};
\addplot[index of colormap=4 of Paired,name path=plus,draw=none] table [x=timesteps, y=rewards_mean_plus_std, col sep=comma]{qrdqn/data/qrdqn_score.csv};
\addplot[index of colormap=4 of Paired,name path=minus,draw=none] table [x=timesteps, y=rewards_mean_minus_std, col sep=comma]{qrdqn/data/qrdqn_score.csv};
\addplot[index of colormap=4 of Paired,fill opacity=0.15] fill between[of=plus and minus];

\end{groupplot}
\node[below=0.1cm, anchor=north west] at (current bounding box.north east) {\pgfplotslegendfromname{exp}};

\end{tikzpicture}
\caption{Crafter scores and rewards with QR-DQN algorithm.}
\label{fig:qrdqn_score}
\end{figure}
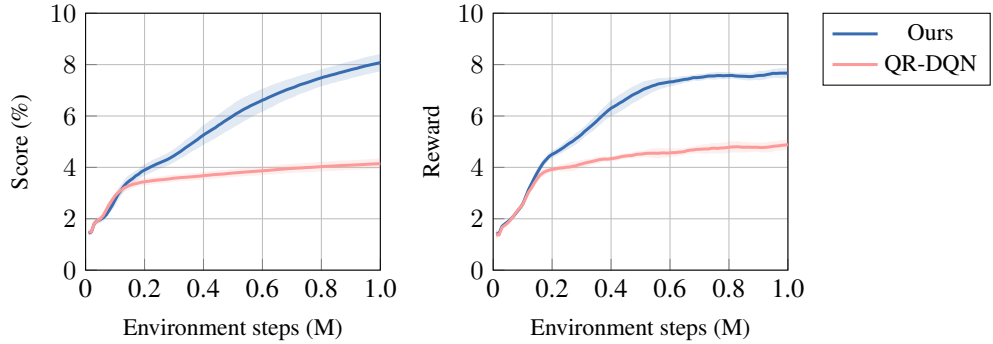

Our implementation of QR-DQN is derived from an open-source implementation (\url{https://github.com/Kaixhin/Rainbow}) of Rainbow \citep{hessel2018rainbow}. We use the original hyperparameter settings for training. For the network architecture, we employ the ResNet encoder, consistent with our PPO implementation. We apply our contrastive learning method to the Q-network encoder prior to each target Q-network update. The default values for the hyperparameters are shown in \Cref{tab:qrdqn_hp}. Our method significantly improves the performance of QR-DQN, as shown in \Cref{fig:qrdqn_score}.

\begin{table}[h]
\centering
\caption{QR-DQN hyperparameters.}
\begin{tabular}{cc}
\toprule
Hyperparameter & Value \\
\midrule
Discount factor & 0.95 \\
Batch size & 64 \\
Replay frequency & 4 \\
Target update & 8000 \\ 
Learning rate & 6.25e-5 \\
Adam $\epsilon$ & 1.5e-4 \\
Max grad norm & 10 \\
Exploration strategy & $\epsilon$-greedy \\
\bottomrule
\end{tabular}
\label{tab:qrdqn_hp}
\end{table}

\section{Application to other environments}

\input{figures/procgen_minigrid_overview}

\subsection{Procgen Heist}

Heist is a procedurally generated door-key environment, whose goal is to steal a gem after unlocking a sequence of blue, green, and red locks, as illustrated in \Cref{fig:procgen_overview}. To open each lock, an agent must collect a key with the corresponding color. We consider unlocking each lock and stealing a gem as achievements. To ensure closer alignment with Crafter, we slightly adjust the reward structure so that an agent receives a reward of 2 for opening each lock and a reward of 10 for successfully stealing a gem. We train the agent in the ``hard'' difficulty mode for 25M environment steps and evaluate its performance in terms of the success rate of gem pilfering and the episode reward. \Cref{fig:procgen_score} shows that our method outperforms PPO by a significant margin throughout training.

\subsection{MiniGrid}

\begin{figure}[ht]
\centering
\begin{tikzpicture}
\begin{groupplot}[
group style = {group size = 2 by 1, horizontal sep=1.5cm},
width=5.5cm,
height=5.0cm,
grid=major,
xmin=0,
xmax=1e6,
scaled x ticks = false,
tick pos = left,
xtick={0, 2e5, 4e5, 6e5, 8e5, 10e5},
xticklabels={0, 0.2, 0.4, 0.6, 0.8, 1.0},
xlabel={Environment steps (M)},
ylabel near ticks,
xlabel style={font=\small},
ylabel style={font=\small},
]
\nextgroupplot[
ylabel=Score (\%),
no markers,
legend style={font=\small,legend columns=1},
legend to name=exp,
ytick={0, 2, 4, 6, 8, 10},
ymin=0.0,
ymax=10,
]

\addlegendimage{index of colormap=4 of Accent,line width=0.4mm}
\addlegendentry{Ours};

\addlegendimage{index of colormap=4 of Paired,line width=0.4mm}
\addlegendentry{PPO};

\addplot[index of colormap=4 of Accent,line width=0.4mm] table [x=timesteps, y=scores_mean, col sep=comma]{minigrid/data/ours_score.csv};
\addplot[index of colormap=4 of Accent,name path=plus,draw=none] table [x=timesteps, y=scores_mean_plus_std, col sep=comma]{minigrid/data/ours_score.csv};
\addplot[index of colormap=4 of Accent,name path=minus,draw=none] table [x=timesteps, y=scores_mean_minus_std, col sep=comma]{minigrid/data/ours_score.csv};
\addplot[index of colormap=4 of Accent,fill opacity=0.15] fill between[of=plus and minus];

\addplot[index of colormap=4 of Paired,line width=0.4mm] table [x=timesteps, y=scores_mean, col sep=comma]{minigrid/data/ppo_large_score.csv};
\addplot[index of colormap=4 of Paired,name path=plus,draw=none] table [x=timesteps, y=scores_mean_plus_std, col sep=comma]{minigrid/data/ppo_large_score.csv};
\addplot[index of colormap=4 of Paired,name path=minus,draw=none] table [x=timesteps, y=scores_mean_minus_std, col sep=comma]{minigrid/data/ppo_large_score.csv};
\addplot[index of colormap=4 of Paired,fill opacity=0.15] fill between[of=plus and minus];

\nextgroupplot[
ylabel=Reward,
no markers,
ymin=0,
ymax=8,
]

\addplot[index of colormap=4 of Accent,line width=0.4mm] table [x=timesteps, y=rewards_mean, col sep=comma]{minigrid/data/ours_score.csv};
\addplot[index of colormap=4 of Accent,name path=plus,draw=none] table [x=timesteps, y=rewards_mean_plus_std, col sep=comma]{minigrid/data/ours_score.csv};
\addplot[index of colormap=4 of Accent,name path=minus,draw=none] table [x=timesteps, y=rewards_mean_minus_std, col sep=comma]{minigrid/data/ours_score.csv};
\addplot[index of colormap=4 of Accent,fill opacity=0.15] fill between[of=plus and minus];

\addplot[index of colormap=4 of Paired,line width=0.4mm] table [x=timesteps, y=rewards_mean, col sep=comma]{minigrid/data/ppo_large_score.csv};
\addplot[index of colormap=4 of Paired,name path=plus,draw=none] table [x=timesteps, y=rewards_mean_plus_std, col sep=comma]{minigrid/data/ppo_large_score.csv};
\addplot[index of colormap=4 of Paired,name path=minus,draw=none] table [x=timesteps, y=rewards_mean_minus_std, col sep=comma]{minigrid/data/ppo_large_score.csv};
\addplot[index of colormap=4 of Paired,fill opacity=0.15] fill between[of=plus and minus];

\end{groupplot}
\node[below=0.1cm, anchor=north west] at (current bounding box.north east) {\pgfplotslegendfromname{exp}};

\end{tikzpicture}
\caption{Minigrid score and reward curves.}
\label{fig:minigrid_score}
\end{figure}
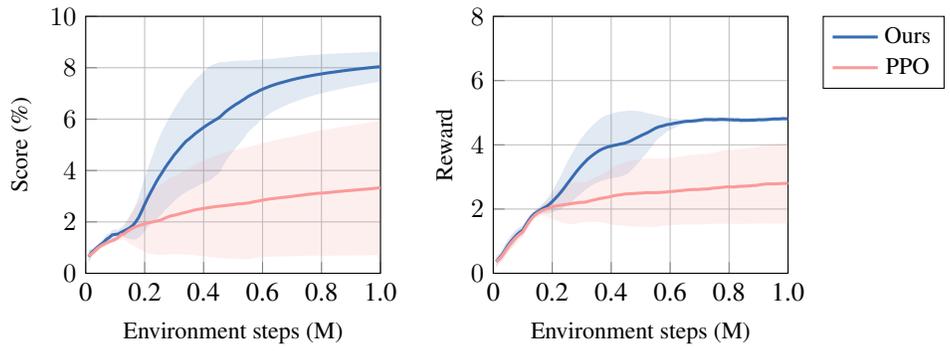

The design of the custom MiniGrid environment takes inspiration by TreeMaze proposed in SEA \citep{zhou2023learning}. An agent must sequentially unlock doors, find keys, and finally reach the green square, as depicted in \Cref{fig:minigrid_overview}. The environment comprises a total of 10 achievements. An agent receives a reward of 1 for unlocking a new achievement, mirroring the reward structure in Crafter. We train an agent for 1M environment steps and evaluate its performance in terms of the geometric mean of success rates and the episode reward, following the same protocol as Crafter. \Cref{fig:minigrid_score} demonstrates that our method outperforms PPO and showcases reduced variance across various training seeds.

\newpage

\begin{figure}[ht]
\centering
\begin{tikzpicture}
\begin{groupplot}[
group style = {group size = 2 by 1, horizontal sep=1.5cm},
width=5.5cm,
height=5.0cm,
grid=major,
xmin=0,
xmax=25e6,
scaled x ticks = false,
tick pos = left,
xtick={0, 5e6, 10e6, 15e6, 20e6, 25e6},
xticklabels={0, 5, 10, 15, 20, 25},
xlabel={Environment steps (M)},
ylabel near ticks,
xlabel style={font=\small},
ylabel style={font=\small},
]
\nextgroupplot[
ylabel=Success rate (\%),
no markers,
legend style={font=\small,legend columns=1},
legend to name=exp,
ytick={0, 0.2, 0.4, 0.6, 0.8, 1.0},
yticklabels={0, 20, 40, 60, 80, 100},
ymin=-0.0,
ymax=1.0,
]

\addlegendimage{index of colormap=4 of Accent,line width=0.4mm}
\addlegendentry{Ours};

\addlegendimage{index of colormap=4 of Paired,line width=0.4mm}
\addlegendentry{PPO};

\addplot[index of colormap=4 of Accent,line width=0.4mm] table [x=step, y=success_mean, col sep=comma]{procgen/data/ppo_infonce_v1_large-hard-success.csv};
\addplot[index of colormap=4 of Accent,name path=plus,draw=none] table [x=step, y=success_mean_plus_std, col sep=comma]{procgen/data/ppo_infonce_v1_large-hard-success.csv};
\addplot[index of colormap=4 of Accent,name path=minus,draw=none] table [x=step, y=success_mean_minus_std, col sep=comma]{procgen/data/ppo_infonce_v1_large-hard-success.csv};
\addplot[index of colormap=4 of Accent,fill opacity=0.15] fill between[of=plus and minus];

\addplot[index of colormap=4 of Paired,line width=0.4mm] table [x=step, y=success_mean, col sep=comma]{procgen/data/ppo_large-hard-success.csv};
\addplot[index of colormap=4 of Paired,name path=plus,draw=none] table [x=step, y=success_mean_plus_std, col sep=comma]{procgen/data/ppo_large-hard-success.csv};
\addplot[index of colormap=4 of Paired,name path=minus,draw=none] table [x=step, y=success_mean_minus_std, col sep=comma]{procgen/data/ppo_large-hard-success.csv};
\addplot[index of colormap=4 of Paired,fill opacity=0.15] fill between[of=plus and minus];

\nextgroupplot[
ylabel=Reward,
no markers,
ymin=0,
ymax=16,
ytick={0, 4, 8, 12, 16},
]

\addplot[index of colormap=4 of Accent,line width=0.4mm] table [x=step, y=reward_mean, col sep=comma]{procgen/data/ppo_infonce_v1_large-hard-reward.csv};
\addplot[index of colormap=4 of Accent,name path=plus,draw=none] table [x=step, y=reward_mean_plus_std, col sep=comma]{procgen/data/ppo_infonce_v1_large-hard-reward.csv};
\addplot[index of colormap=4 of Accent,name path=minus,draw=none] table [x=step, y=reward_mean_minus_std, col sep=comma]{procgen/data/ppo_infonce_v1_large-hard-reward.csv};
\addplot[index of colormap=4 of Accent,fill opacity=0.15] fill between[of=plus and minus];

\addplot[index of colormap=4 of Paired,line width=0.4mm] table [x=step, y=reward_mean, col sep=comma]{procgen/data/ppo_large-hard-reward.csv};
\addplot[index of colormap=4 of Paired,name path=plus,draw=none] table [x=step, y=reward_mean_plus_std, col sep=comma]{procgen/data/ppo_large-hard-reward.csv};
\addplot[index of colormap=4 of Paired,name path=minus,draw=none] table [x=step, y=reward_mean_minus_std, col sep=comma]{procgen/data/ppo_large-hard-reward.csv};
\addplot[index of colormap=4 of Paired,fill opacity=0.15] fill between[of=plus and minus];

\end{groupplot}
\node[below=0.1cm, anchor=north west] at (current bounding box.north east) {\pgfplotslegendfromname{exp}};

\end{tikzpicture}
\caption{Procgen Heist success rate and reward curves.}
\label{fig:procgen_score}
\end{figure}
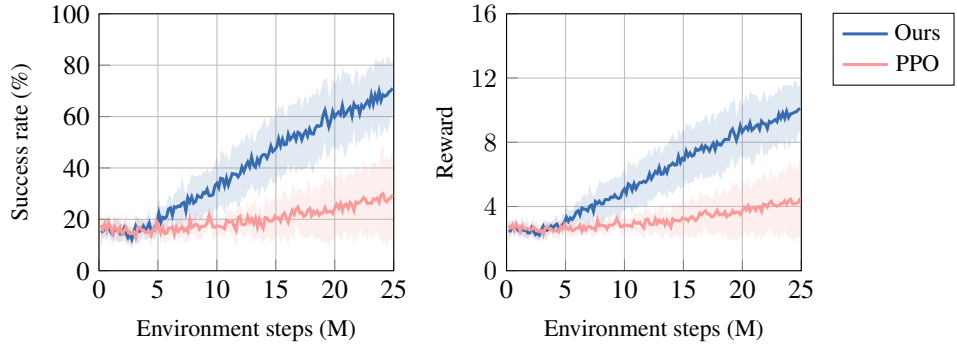

\end{document}